\documentclass[letterpaper]{article} % DO NOT CHANGE THIS
\PassOptionsToPackage{x11names, dvipsnames, table, xcdraw}{xcolor}
\usepackage{aaai2026}  % DO NOT CHANGE THIS
\usepackage{times}  % DO NOT CHANGE THIS
\usepackage{helvet}  % DO NOT CHANGE THIS
\usepackage{courier}  % DO NOT CHANGE THIS
\usepackage[hyphens]{url}  % DO NOT CHANGE THIS
\usepackage{graphicx} % DO NOT CHANGE THIS
\usepackage{natbib}  % DO NOT CHANGE THIS AND DO NOT ADD ANY OPTIONS TO IT
\usepackage{caption} % DO NOT CHANGE THIS AND DO NOT ADD ANY OPTIONS TO IT
\usepackage{algorithm}
\usepackage{algorithmic}

\usepackage{newfloat}
\usepackage{listings}
\DeclareCaptionStyle{ruled}{labelfont=normalfont,labelsep=colon,strut=off} % DO NOT CHANGE THIS
\floatstyle{ruled}
\newfloat{listing}{tb}{lst}{}
\floatname{listing}{Listing}
\usepackage{enumitem}
\usepackage{algorithm}
\usepackage{algorithmic}
\usepackage{booktabs}
\usepackage{multirow}
\usepackage{arydshln}
\usepackage{pgfplots}
\usepackage[most]{tcolorbox}    % 提供 tcolorbox 环境及其增强功能
\usepackage{amsmath}            % 支持数学公式（如 $6^3$）
\usepackage{amssymb}            % 提供更多数学符号
\usepackage{cleveref}
\usepackage{graphicx}           % 如果将来要插入图片（非必须）
\usepackage{listings}
\usepackage{siunitx}            % 数字列对齐
\usepackage{makecell}
\usepackage{tabularx}
\usepackage{forest}
\usepackage{fancyvrb}           % 更好的 verbatim 支持
\usepackage{changepage}         % 提供 adjustwidth 环境
\usepackage{xcolor}             % 选项在最上方配置中加
\usepackage{xspace}
\usepackage{placeins}
\usepackage{pifont}
\usepackage{booktabs, multirow, colortbl, xcolor}  % 放在导言区

\pgfplotsset{compat=1.17}
\title{Reasoning or Memorization? Unreliable Results of Reinforcement Learning\\Due to Data Contamination}
\author{
    Mingqi Wu\textsuperscript{\rm 1}\equalcontrib, 
    Zhihao Zhang\textsuperscript{\rm 1 \rm 2}\equalcontrib, 
    Qiaole Dong\textsuperscript{\rm 1}\equalcontrib,\\
    Zhiheng Xi\textsuperscript{\rm 1}, 
    Jun Zhao\textsuperscript{\rm 1}, 
    Senjie Jin\textsuperscript{\rm 1}, 
    Xiaoran Fan\textsuperscript{\rm 1}, 
    Yuhao Zhou\textsuperscript{\rm 1}, 
    Huijie Lv\textsuperscript{\rm 1 \rm 2}, 
    Ming Zhang\textsuperscript{\rm 1}, 
    Yanwei Fu\textsuperscript{\rm 1}, 
    Qin Liu\textsuperscript{\rm 4}, 
    Songyang Zhang\textsuperscript{\rm 2}, 
    Qi Zhang\textsuperscript{\rm 1 \rm 2 \rm 3}\thanks{Corresponding author.}
}
\affiliations{
    \textsuperscript{\rm 1}Fudan University\\
    \textsuperscript{\rm 2}Shanghai Artificial Intelligence Laboratory\\
    \textsuperscript{\rm 3}Shanghai Key Lab of Intelligent Information Processing\\
    \textsuperscript{\rm 4}University of California, Davis\\
    
    {qz}@fudan.edu.cn, {qinli}@ucdavis.edu
}

\usepackage{bibentry}
\newcommand{\qwenSevenB}             {Qwen2.5-7B\xspace}
\newcommand{\qwenSevenBInstruct}     {Qwen2.5-7B-Instruct\xspace}
\newcommand{\qwenmathSevenB}         {Qwen2.5-Math-7B\xspace}
\newcommand{\qwenmathSevenBInstruct} {Qwen2.5-Math-7B-Instruct\xspace}
\newcommand{\llamaEightB}            {Llama3.1-8B\xspace}
\newcommand{\llamaEightBInstruct}    {Llama3.1-8B-Instruct\xspace}

\newcommand{\qwenThreeFourB}         {Qwen3-4B\xspace}
\newcommand{\qwenThreeFourBBase}     {Qwen3-4B-Base\xspace}
\newcommand{\qwenThreeEightB}        {Qwen3-8B\xspace}
\newcommand{\qwenThreeEightBBase}    {Qwen3-8B-Base\xspace}
\newcommand{\qwenThreeFourteenB}     {Qwen3-14B\xspace}
\newcommand{\qwenThreeFourteenBBase} {Qwen3-14B-Base\xspace}

\newcommand{\randomcalculation}{RandomCalculation\xspace}
\newcommand{\greedyonly}{Greedy (w/o Template)\xspace}
\newcommand{\sampleonly}{Avg@16 (w/o Template)\xspace}
\newcommand{\templategreedy}{Greedy (w/ Template)\xspace}
\newcommand{\templatesample}{Avg@16 (w/ Template)\xspace}

\begin{document}

\maketitle

\begin{abstract}
Reasoning in large language models has long been a central research focus, and recent studies employing reinforcement learning (RL) have introduced diverse methods that yield substantial performance gains with minimal or even no external supervision. Surprisingly, some studies even suggest that random or incorrect reward signals can enhance performance. However, these breakthroughs are predominantly observed for the mathematically strong Qwen2.5 series on benchmarks such as MATH-500, AMC, and AIME, and seldom transfer to models like Llama, which warrants a more in-depth investigation. In this work, our empirical analysis reveals that pre-training on massive web-scale corpora leaves Qwen2.5 susceptible to data contamination in widely used benchmarks. Consequently, conclusions derived from contaminated benchmarks on Qwen2.5 series may be unreliable. To obtain trustworthy evaluation results, we introduce a generator that creates fully clean arithmetic problems of arbitrary length and difficulty, dubbed \textit{RandomCalculation}. Using this leakage-free dataset, we show that only accurate reward signals yield steady improvements that surpass the base model’s performance boundary in mathematical reasoning, whereas random or incorrect rewards do not. 
Moreover, we conduct more fine-grained analyses to elucidate the factors underlying the different performance observed on the MATH-500 and \randomcalculation benchmarks. 
Consequently, we recommend that future studies evaluate models on uncontaminated benchmarks and, when feasible, test various model series to ensure trustworthy conclusions about RL and related methods.

\end{abstract}

\begin{links}
    \textbf{Code}: https://github.com/wumingqi/LLM-Math-Evaluation
    % \link{Datasets}{https://aaai.org/example/datasets}
    % \link{Extended version}{https://arxiv.org/abs/2507.10532}
\end{links}

\section{Introduction}
In recent years, advances in reinforcement learning (RL) have markedly strengthened the reasoning abilities of large language models (LLMs). Flagship systems, including OpenAI’s o1~\citep{openai-o1, openai-o3, openai-gpt4o}, DeepSeek-R1~\citep{DeepSeek-R1}, and QwQ~\citep{qwq,qwq32b}, already match or exceed human-level accuracy on a variety of challenging benchmarks. Among open-source contenders that excel in these mathematical benchmarks, the Qwen family~\citep{DBLP:qwen2.5-tech-report, DBLP:qwen2.5-math-tech-report, DBLP:Qwen3-Technical-Report}, ranging from 0.5B to 72B and pre-trained on up to 36T high-quality tokens, produces state-of-the-art results in language understanding, mathematics, programming, and preference alignment.

Within this landscape, mathematical reasoning emerges as a particularly discriminative test bed because it demands precise symbolic manipulation and multi-step logical deduction. Standard suites such as MATH-500~\citep{DBLP:conf/nips/HendrycksBKABTS21}, AIME~\citep{numina_math}, AMC~\citep{numina_math}, and Minerva Math~\citep{DBLP:minerva_math} require models to parse natural-language problem statements, uncover the latent mathematical structure, and generate exact numeric answers. Recent work has further enhanced this capability by reinforcement learning with verifiable rewards (RLVR)~\citep{DeepSeek-R1}: rule-based reward that returns 1 when the predicted answer equals ground truth and 0 otherwise. Because the reward is computed analytically, RLVR removes the need for a separate learned reward model, lowering computational cost while providing a precise training signal, especially attractive for domains like mathematics, where solutions are unambiguous.

%%%%%%%%%%%%%%%%%%%%%%%%%%%%%%%%%%%%%%%%%%%%%%%%%%%%%%%%%%%%%%%%%%%%%%%%%%%%%%%%
%
% 【图】MATH-500在Qwen上面的泄露情况示例，带有Llama的输出作为对比
% \input{figure/figure_math500_qwen_and_llama_output_comparison}

%
% 图表：对比展示两个模型的输出
%

% 统一设置紧凑的tcolorbox样式变量
\tcbset{
  mytightbox/.style={
    boxsep=1mm,
    left=1mm,
    right=1mm,
    top=1mm,
    bottom=1mm
  }
}

\begin{figure}[!ht]
\small
\centering
\begin{minipage}{\linewidth}
% \begin{tcolorbox}[
%   enhanced,
%   attach boxed title to top center={yshift=-3mm},
%   colbacktitle=red!75!black,
%   colframe=gray!75!black,
%   colback=white,
%   width=\textwidth,
%   boxed title style={size=small, colback=red!75!black},
%   fonttitle=\bfseries\color{white},
%   title={}, %此处可以改标题
% ]

% 顶部框：题目部分
\begin{tcolorbox}[
  mytightbox,
  colback=gray!4!white,
  colframe=gray!85!black,
  coltitle=white,
  title=\textbf{Example from MATH-500},
  fonttitle=\bfseries,
]
\textbf{Problem}: For how many positive integers $n>1$ is it true that $2^{24}$ is a perfect $n^{\text{th}}$ power? 

\textbf{Answer}: \boxed{7}

\end{tcolorbox}

\begin{tcolorbox}[
  mytightbox,
  colback=gray!4!white,
  colframe=gray!85!black,
  title=\textbf{Partial Prompt},
  fonttitle=\bfseries,
]
\textbf{Prompt}: For how many positive integers $n>1$ is
\end{tcolorbox}

% 下方左右并排等高的两个框
\noindent
\begin{minipage}[t]{0.495\textwidth}
  \begin{tcolorbox}[
    mytightbox,
    % colback=cyan!10, 
    % colframe=SkyBlue!80!black, 
    % coltitle=white,
    colback=teal!2!white, 
    colframe=teal!80!black, 
    coltitle=white,
    title=\textbf{Qwen2.5-7B Output},
    fonttitle=\bfseries,
    equal height group=mathcompare,
    height fixed for=all,
  ]

{\bfseries it true that $2^{24}$ is a perfect $n^{\text{th}}$ power?}
 
% \vspace{1\baselineskip}

To determine for how many positive integers \( n > 1 \) the number \( 2^{24} \) is a perfect \( n \)-th power...

% First, we need to find...

% ...

Let's count these values. There are \( 7 \) such values...

% ...

The final answer is \(\boxed{7}\).

  \end{tcolorbox}
\end{minipage}
\hfill
\begin{minipage}[t]{0.495\textwidth}
  \begin{tcolorbox}[
    mytightbox,
    colback=teal!2!white, 
    colframe=teal!80!black, 
    coltitle=white,
    title=\textbf{Llama3.1-8B Output},
    fonttitle=\bfseries,
    equal height group=mathcompare,
    height fixed for=all,
  ]
the set of positive divisors of $n$ (including $1$ and $n$) less than $50$ ?

\noindent
\textbf{(A)} 15 
\textbf{(B)} 16 
\textbf{(C)} 17 ...
%\textbf{(D)} 18 
%\textbf{(E)} 19

% \vspace{1\baselineskip}

The number of positive divisors of $n$ is given by...

% ...

The number of factors of each of these numbers is...

% ...

The answer is $\boxed{\textbf{(B) }16}$ 

  \end{tcolorbox}
\end{minipage}

% \end{tcolorbox}
\end{minipage}
%
% Qwen2.5模型在MATH-500上背诵痕迹示例。这里我们使用题目的40%比例作为prompt。Qwen2.5-7B不仅精确地补全了原来的题目，而且给出了完整的推理步骤和正确的答案。Llama3.1-8B则给出了错误的补全和答案。
%
\caption{Memorization Example of Qwen2.5 on MATH-500.
In this case, the first 40\% of the original problem is used as the prompt, and the generation is performed under the \textbf{\textit{\greedyonly}} configuration (see Table~\ref{tab:gen-configs}). The \qwenSevenB model accurately reproduces the original question verbatim and, moreover, generates a complete and precise chain of reasoning that yields the correct answer. In contrast, Llama3.1-8B produces an incorrect completion and ultimately arrives at an incorrect answer.}
\label{fig:qwen_and_llama_output_on_math_500_example}
\end{figure}

%
%%%%%%%%%%%%%%%%%%%%%%%%%%%%%%%%%%%%%%%%%%%%%%%%%%%%%%%%%%%%%%%%%%%%%%%%%%%%%%%%

Although RL nominally depends on accurate reward signals to guide training, recent studies~\citep{TTRL, Spurious-Rewards} find that even random or incorrect rewards can improve Qwen's performance on standard math benchmarks, while the same procedures offer little or no benefit to \llamaEightB~\citep{llama3}. To understand why these seemingly problem-agnostic rewards help Qwen but not Llama, we undertake a systematic comparison of two model families under identical training protocols. We consider two working hypotheses to explain this phenomenon. (i) \textbf{Data Contamination}: Considering Qwen2.5 is pre-trained on massive web-scale corpora from the Internet, including GitHub repositories that store benchmark problems alongside their official solutions. If segments of the evaluation benchmarks leaked into the pre-training corpus, spurious rewards could cue the model to retrieve memorized answers rather than acquire new reasoning skills. (ii) \textbf{Strong Math Capacity}: Qwen’s pre-training endows it with better mathematical capacity than Llama, so even noisy policy-gradient updates appear to help on MATH-500. However, if strong capacity is the real driver, the same spurious rewards should still work on a clean benchmark. Distinguishing between these possibilities requires both a leakage audit and a rigorously out-of-distribution RLVR evaluation.
% , which we present in Section~\ref{sec:memorization-math500} and Section~\ref{sec:rlvr-random-calculation}.

To assess the extent of potential data contamination in popular mathematical benchmarks, we propose two metrics: \textbf{partial-prompt completion rate} (can the model reconstruct the tail of a problem?) and \textbf{partial-prompt answer accuracy} (can the model give the correct answer with an incomplete problem?). As shown in Fig.~\ref{fig:qwen_and_llama_output_on_math_500_example}, Qwen can indeed complete the problem accurately and provide the correct answer, whereas Llama fails. Furthermore, prompting with the first 60\% of each MATH-500 problem, Qwen2.5-Math-7B regenerates the remaining 40\% with a 54.60\% exact-match rate and answers 53.6\% of these incomplete problems correctly. In contrast, \llamaEightB scores 3.8\% and 2.4\% on both metrics. Crucially, on the newly released LiveMathBench~(version 202505)~\citep{DBLP:LiveMathBench}, 
% , a post-hoc benchmark compiled after the release of the Qwen2.5, 
Qwen's completion rate drops sharply to 0.0\%, consistent with Llama's 0.0\%. Its partial-prompt answer accuracy also falls to just 2.0\%, comparable to Llama's 1.0\%. 
These results confirm that Qwen's pre-training corpus suffers from test data contamination.
% These results confirm that the earlier gains on MATH-500 may stem from memorized content rather than genuine reasoning. 
So, results derived from MATH-500 and similar datasets for Qwen should be interpreted with caution.

Based on this, we attribute that data contamination is the main factor behind the `magical' success of spurious rewards on Qwen. To test this claim, we first create a clean benchmark (\emph{i.e.}, \randomcalculation, example shown in Fig.~\ref{fig:example_random_calculation}): We use automatic generator to construct arithmetic expressions of arbitrary length with random operands and operators, guaranteeing that every instance post-dates the public release of Qwen. Zero-shot evaluation on this benchmark shows no memorization: the accuracy of Qwen2.5 declines monotonically with the number of computation steps. To isolate the effect of rewards, we next trained {\qwenmathSevenB} under the standard RLVR protocol on two subsets. The outcome is unambiguous:
Correct rewards deliver consistent performance gains, surpassing the model’s performance ceiling. In contrast, random rewards make training highly unstable, yielding no reliable improvement, while inverse rewards rapidly erode the model’s mathematical-reasoning ability. These results rule out the `Strong Math Capacity' hypothesis and directly imply `Data Contamination': once leakage is removed, spurious gains evaporate.

To further test this hypothesis, we measure the similarity between model outputs before and after RL. In MATH-500, the pre- and post-RL responses exhibited substantially higher lexical overlap than in \randomcalculation, indicating that Qwen inadvertently retrieves its memory and answers during RL with spurious rewards. A more detailed token-level analysis also supports this: the token-level KL divergence between the pre- and post-RL models is significantly lower for MATH-500. These results strengthen our hypothesis that data contamination leads to successful RL through spurious rewards on Qwen series. Based on these findings, we recommend that future work should test on uncontaminated benchmarks or more diverse model series to draw trustworthy conclusions about RL-related methods.

Contributions of our work can be summarized as follows:
\begin{itemize}
    \item We conduct a systematic leakage audit of math benchmarks with two novel metrics and demonstrate that Qwen suffers from data contamination on public benchmarks.
    % ’s sudden performance surge on MATH-500 under spurious-reward is chiefly attributable to unfair data contamination and memorization, rather than its strong math capacity.
    \item We propose an automatic generator that creates arbitrarily long arithmetic expressions.
    % and a corresponding out-of-distribution evaluation protocol. 
    Zero-shot evaluation on this dataset exposes the absence of memorization, 
    % and places Qwen and Llama on equal footing
    enabling fair assessment of RL methods.
    \item Using this clean dataset, we conduct RL experiments and demonstrate that only \emph{correct} reward yields stable improvement, whereas spurious rewards provide no benefit.
    % , thereby highlighting the central role of reward fidelity. 
    \item We reveal that spurious rewards solely enable the retrieval of memory from pre-training, leading to spurious performance improvement on MATH-500.
    % on MATH-500, Qwen’s pre- and post-training outputs display an even higher degree of textual overlap, accompanied by a notably more stable full-vocabulary probability distribution.
\end{itemize}

\section{Related Works}
\subsection{RL on Qwen2.5 for Mathematical Reasoning}
A growing body of work investigates how reinforcement learning (RL) can amplify the mathematical-reasoning capacity of the open-source Qwen2.5 family.  Early studies use verifiable rewards that score an answer as 1 / 0 by exact numerical agreement.  Test-time RL~\citep{TTRL} applies this signal on-the-fly during inference and yields sizeable gains on MATH-500 and AIME2024.  Subsequent efforts pursue extreme data efficiency: few labeled~\citep{1-shot-RL, Few-Shot-RL} or even unlabeled examples~\citep{one-shot-EM, Absolute-Zero} can suffice to boost performance.
A parallel research replaces external supervision with \emph{intrinsic} signals derived from the model itself: \citet{RENT, EM-RL} reward low-entropy output distributions, while \citet{INTUITOR, Self-Train} rely on self-consistency or self-certainty as feedback. These approaches report large jumps on Qwen2.5-Math-7B, occasionally matching or surpassing stronger supervised baseline. Other variants explore noisy~\citep{lv-PRP} or even random rewards~\citep{Spurious-Rewards} for Qwen2.5. However, these methods fails to transfer to Llama or OLMo~\citep{DBLP:olmo}, suggesting model-specific idiosyncrasies.
% The generality of these gains, however, has been questioned.  
% \citet{NoFreeLunch} observe an `underconfidence-to-overconfidence' trajectory in RLIF (Reinforcement Learning from Internal Feedback): accuracy rises early but later collapses as policy entropy shrinks. 
% \citet{Spurious-Rewards} further shows that  can still improve  yet fail   
% Together, these findings suggest that while RL can unlock strong mathematical competence in Qwen2.5, its effectiveness depends on careful reward design, entropy control, and the underlying model's capacity.

\subsection{Factors Influencing Performance on Math}
The choice of pretraining corpora plays a crucial role in shaping the reasoning abilities of LLMs, particularly in mathematical domains. Several math-specific datasets~\citep{OpenWebMath, InfiMM-WebMath, MathPile, SmolLM2} have been proposed and shown to significantly enhance performance on relevant benchmarks. Further, \citet{OctoThinker} finds that mid-training Llama models on high-quality mathematical corpora substantially improve their capacity, both at the base level and after reinforcement learning.
However, evaluation of math capability can be misleading due to potential test data contamination. \citet{DBLP:Benchmarking-Benchmark} observed that certain widely-used benchmarks may be partially included in the pretraining corpus of LLMs, including early versions of Qwen.  On the other hand, \citet{DBLP:Understanding-R1} found that omitting dialogue-style prompting can lead to improved mathematical reasoning. These observations underscore the presence of multiple confounding factors in evaluating model performance, motivating the need for rigorous and systematic analysis.

%%%%%%%%%%%%%%%%%%%%%%%%%%%%%%%%%%%%%%%%%%%%%%%%%%%%%%%%%%%%%%%%%%%%%%%%%%%%%%%%
% 【图】RandomCalculation数据集示例
%
% \input{figure/figure_five_step_calculation_example}
%
% 自定义随机计算数据集示例
%

% 统一设置紧凑的tcolorbox样式变量
\tcbset{
  mytightbox/.style={
    boxsep=1mm,
    left=1mm,
    right=1mm,
    top=1mm,
    bottom=1mm
  }
}
\begin{figure}
\small
\centering
\begin{tcolorbox}[
  mytightbox,
  colframe=DodgerBlue4, %
  colback=white,
  coltitle=white,
  title=\textbf{10-Step Calculation},
  fonttitle=\bfseries
]
\textbf{Problem}: Evaluate this LaTeX numerical expression step-by-step and give the final value within \textbackslash boxed\{\}: 

\[
\frac{94}{2} + \left( \frac{73^2 \cdot (62 - 10)}{\left( \frac{\frac{65}{9} + 47}{\frac{\frac{49}{7} \cdot 81}{62^2}} \right)} \right) \cdot \left( \frac{41}{6} + \frac{12}{7} \right)
\]

\textbf{Answer}: \boxed{6490.42220471333}

\end{tcolorbox}

\caption{Examples of \textit{{\randomcalculation}} dataset.}
\label{fig:example_random_calculation}
\end{figure}

%
%%%%%%%%%%%%%%%%%%%%%%%%%%%%%%%%%%%%%%%%%%%%%%%%%%%%%%%%%%%%%%%%%%%%%%%%%%%%%%%%

\section{Experimental Setup}

\subsection{Model Selection}
Prior researches on mathematical reasoning with LLMs focus predominantly on the Qwen-2.5~\citep{DBLP:qwen2.5-tech-report, DBLP:qwen2.5-math-tech-report}. Accordingly, we center our study on four representative checkpoints from this series: \qwenSevenB, \qwenSevenBInstruct, \qwenmathSevenB, and \qwenmathSevenBInstruct. For a controlled comparison, we also evaluate \llamaEightB and \llamaEightBInstruct~\citep{llama3}, which possess comparable parameter counts and thus help isolate model-specific differences in behavior.

\subsection{Evaluation of Memorization Capability}
\label{sec::Memorization Capability}
We assess the model's memorization of benchmark data, \emph{i.e.}, data contamination, using two metrics: \textbf{Partial-Prompt Completion Rate} and \textbf{Partial-Prompt Answer Accuracy}.

Specifically, we prompt the model to complete the remaining parts of a problem based on partial prefixes. To evaluate its performance, we use the \textbf{Partial-Prompt Completion Rate} measured by ROUGE-L~\citep{lin-2004-rouge}, which calculates the overlap of the longest common subsequence between the generated and reference text, capturing fluency and sentence-level structure. Additionally, we utilize Exact Match (EM) accuracy, which checks if the model’s output exactly matches the reference. The final EM score is the average across all instances. Higher EM indicates a greater proportion of partial-prompts that model can recall exactly.
% The most frequently used variants include:
% \begin{itemize}[leftmargin=1em,itemsep=2pt,topsep=2pt]
%     \item \textbf{ROUGE-N}: Measures n-gram overlap between texts. 
%     % ROUGE-1 and ROUGE-2 refer to unigram and bigram recall respectively.
%     \item \textbf{ROUGE-L}: Measures the longest common subsequence between the generated and reference text.
%     % capturing fluency and sentence-level structure.
%     \item \textbf{ROUGE-Lsum}: A variant of ROUGE-L adapted for multi-sentence summarization evaluation.
% \end{itemize}

% In this work, we utilize \textbf{average ROUGE-L} score to measure the model's ability to reconstruct the remaining parts of a problem based on partial prefixes, serving as an indicator of the model's memorization capacity.
% 2、介绍一下EM指标的计算

% \paragraph{\textbf{Exact Match (EM).}}
% EM is a binary accuracy metric that checks whether the model’s continuation exactly reproduces the reference.  
% For each instance, let \(y\) denote the model-generated continuation and \(y^{*}\) denote the ground-truth continuation.  if $y = y^{*}$, $\text{EM}=1$; otherwise $\text{EM}=0$. We first compute the \textsc{rouge}\,-L score between the \(y\) and the \(y^{*}\), then assign
% \[
% \text{EM}(y,y^{*}) \;=\;
% \begin{cases}
% 1, & \text{if } \textsc{rouge-L}(y,y^{*}) = 1,\\[4pt]
% 0, & \text{otherwise}.
% \end{cases}
% \]

% \subsubsection{Partial-Prompt Answer Accuracy}
% 对于部分的题目prompt，我们让模型自由生成，并最后在生成中进行匹配是否包含了正确答案。以此来检测正确答案在模型补全中的占比情况，进一步反应可能存在的数据污染现象。 
% \paragraph{Answer-Match Accuracy.} 
Besides, for each question, we supply the model with only a truncated prompt (\emph{e.g.}, the first \(60\%\) of the original problem) and allow it to generate an unconstrained continuation.  
After generation, we check whether the completion contains the ground-truth answer; if so, the instance is scored as correct.  
\textbf{Partial-Prompt Answer Accuracy} is defined as the fraction of prompts for which the model's continuation embeds the correct answer.  
A high accuracy indicates that the model frequently `recovers' the answer even from a partial problem, which in turn may signal data contamination.

\subsection{RLVR-Based Evaluation}
% Here we present the GRPO algorithm employed for RLVR in our experiment, and then detail the continuous reward function tailored to our \textit{RandomCalculation} benchmark.
\noindent\textbf{Group Relative Policy Optimization} (GRPO)~\citep{DeepSeekMath} is adopted as our RLVR algorithm. Formally, for each question $q$, GRPO samples a group of outputs $\{o_1, \cdots, o_G\}$  from the old policy  $\pi_{\theta_{\text{old}}}$  and then optimizes the policy model by maximizing the following objective:
\begin{equation}
\begin{aligned}
    \mathcal{J}&_{\text{GRPO}}(\theta) = \mathbb{E}_{q, \{o_i\}_{i=1}^G} \frac{1}{G}\sum_{i=1}^G\frac{1}{|o_i|} \sum_{t=1}^{|o_i|} \left\{ \min \left(r_{i,t} \hat{A}_{i,t}, \right.\right.\\
    & \left.\left.\text{clip} \left( r_{i,t}, 1 - \epsilon, 1 + \epsilon \right)\hat{A}_{i,t} \right) - \beta \mathbb{D}_{\text{KL}}\left[\pi_{\theta} || \pi_{\text{ref}}\right]\right\} ,
\end{aligned}
\label{eq:GRPO-obj}
\end{equation}
where $r_{i,t}=\frac{\pi_\theta(o_{i,t} | q, o_{i,<t})}{\pi_{\theta_{\text{old}}}(o_{i,t} | q, o_{i,<t})}$, $\epsilon$ and $\beta$ are hyperparameters, and $\hat{A}_{i,t}$  is the advantage calculated based on the relative rewards of the outputs inside each group only.

%
% 奖励函数介绍
%

\paragraph{Spurious Reward.}
Following \citet{Spurious-Rewards}, we consider following spurious reward types for RLVR:

\begin{itemize}[leftmargin=1.6em]
    % \item \textbf{Correct}: assigns~1 to correct answer and~0 otherwise.
    \item \textbf{Random}: assigns~\(1\) with probability \(\gamma\) and \(0\) otherwise (\(\gamma = 0.5\) in our experiments).
    \item \textbf{Inverted}: flips the correct signal, \emph{i.e., \(1 - \text{correct}\)}, so that correct solutions receive~0 and incorrect ones~1.
    \item \textbf{Mv-incorrect}: uses the majority-voted incorrect labels from the model, assigning a reward of 1 when the model output matches an incorrect label, and 0 otherwise.
\end{itemize}

\section{Results \& Analysis}
\subsection{Spurious Rewards on MATH-500}
\label{sec:rlvr-math500}
\begin{figure}
  \centering
  \includegraphics[width=0.95\linewidth]{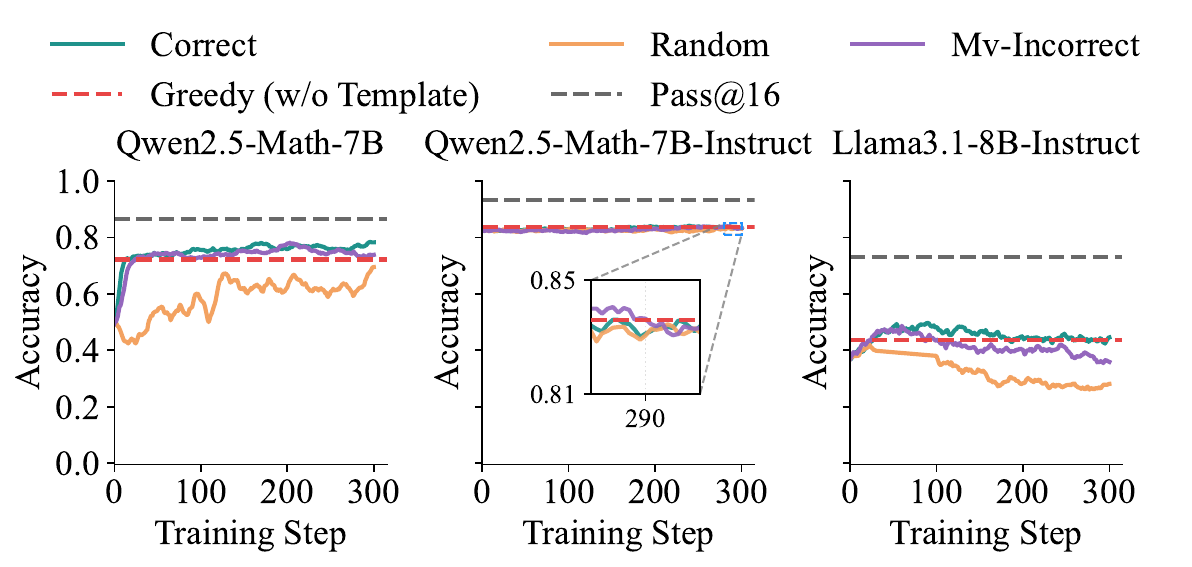}
  % Qwen2.5-Math-7B和Llama3.1-8B-Instruct在不同的奖励信号下进行RLVR训练，在MATH-500数据集上的准确率走势。
  \caption{Accuracy on the \textbf{MATH-500} for \qwenmathSevenB, \qwenmathSevenBInstruct, and \llamaEightBInstruct trained with RLVR under various reward signals. Greedy and pass@16 scores are reported \emph{without} template.}\label{fig:rlvr_qwen_llama_math500}
\end{figure}

Following the work of \citet{Spurious-Rewards}, we replicate the performance of \qwenmathSevenB and \llamaEightBInstruct on the MATH-500 benchmark under various reward signal configurations, using the same experimental setup. The accuracy curves of MATH-500 are shown in Fig.~\ref{fig:rlvr_qwen_llama_math500}. Interestingly, the results demonstrate that while random rewards and mv-incorrect rewards noticeably boost accuracy for \qwenmathSevenB, they have little or even adverse impacts on the performance of \llamaEightBInstruct. Additionally, we apply the same RLVR procedure to \qwenmathSevenBInstruct and discover that the resulting gains are marginal when compared with those of \qwenmathSevenB, indicating that the two Qwen variants exhibit differential sensitivity to RLVR under spurious rewards. 

\begin{figure}
  \centering
  \includegraphics[width=0.95\linewidth]{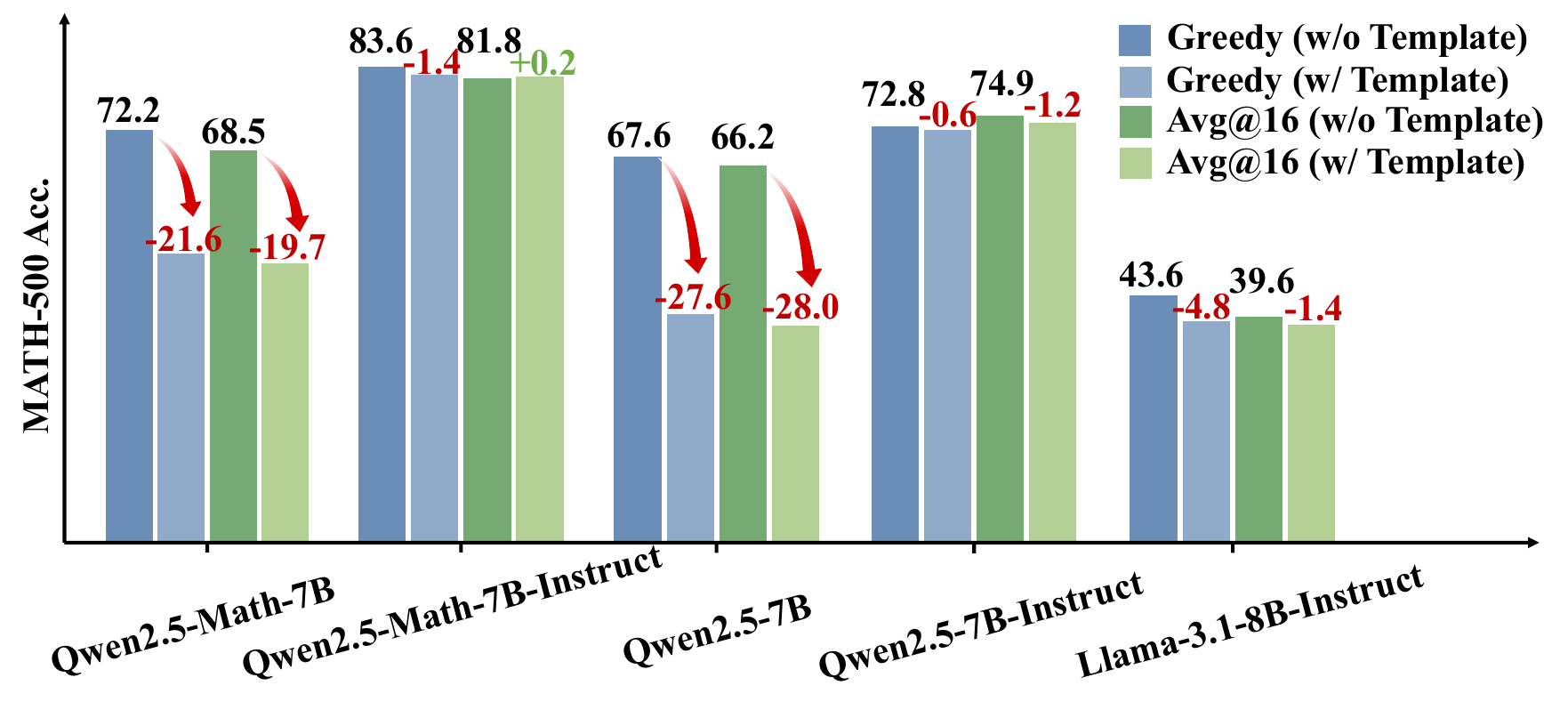}
  \caption{Accuracy (\%) of \textbf{Qwen} and \textbf{Llama} models on the \textbf{MATH-500} dataset under different generation configurations, using original questions as prompts. More detailed results can be found in Tab. 4 of Appendix.}
  % 上方有引用 Tab.~\ref{tab:math-500-comparison}
  \label{fig:math-500-bar-qwen-llama}
\end{figure}

Considering the \textit{base} and \textit{instruct} variants of Qwen are trained under different paradigms: the former is pre-trained as a general language model without exposure to any dialogue-specific templates, whereas the latter undergoes an additional instruction-tuning stage on data wrapped in a fixed dialogue template. This mismatch creates a training–testing gap for the base model at the start point of RLVR, and its initial accuracy is therefore likely underestimated. Consequently, to obtain a fair estimate of each model’s starting point, we next measure performance under four decoding configurations as shown in Tab.~\ref{tab:gen-configs}. The corresponding results are summarized in Fig.~\ref{fig:math-500-bar-qwen-llama}. Surprisingly, we discover that applying the official chat template substantially degrades performance for the Qwen base model: both \qwenSevenB and \qwenmathSevenB suffer pronounced drops once the template is enabled.
% , whether we use greedy or avg@16 sampling. 

%%%%%%%%%%%%%%%%%%%%%%%%%%%%%%%%%%%%%%%%%%%%%%%%%%%%%%%%%%%%%%%%%%%%%%%%%%%%%%%%
%
% 【表】生成配置
% \input{table/generation_configuration}
\begin{table}[!t]
\centering
\small
\setlength{\tabcolsep}{4pt}  % 调小列间距

\begin{tabular}{lcccccc}
\toprule
\textbf{Configuration} & \textbf{\# Sample} & \textbf{Temp.} & \textbf{Top-P} & \textbf{Top-K} \\
\midrule
Greedy (w/o Template)            & 1 & 1.0 & 1.0  & 1 \\
Avg@16 (w/o template)           & 16 & 0.7 & 0.8  & 20 \\
Greedy (w/ Template)     & 1 & 1.0 & 1.0  & 1 & \\
Avg@16 (w/ template)   & 16 & 0.7 & 0.8  & 20 & \\ 
\bottomrule
\end{tabular}

%
% 不同的生成配置使用的vLLM采样参数
%
\caption{The sampling parameters used under different generation configurations. Greedy sampling is performed using the default \texttt{model.generate(\ldots)} function, while random sampling is implemented using \texttt{vLLM}. w/ Template indicates the usage of official chat template.}
% \textit{\textbf{\greedyonly}} refers to greedy decoding with the default \texttt{model.generate(\ldots)} call. \textit{\textbf{\sampleonly}} indicates that sampling 16 candidate answers with \texttt{vLLM} and report their average accuracy. \textit{\textbf{\templategreedy}} denotes the use of the official chat template combined with the default greedy decoding, whereas \textit{\textbf{\templatesample}} applies the official chat template together with the average over 16 rollout samples. In all configurations, the max-tokens is set to 4096.}
\label{tab:gen-configs}
\end{table}
%
%%%%%%%%%%%%%%%%%%%%%%%%%%%%%%%%%%%%%%%%%%%%%%%%%%%%%%%%%%%%%%%%%%%%%%%%%%%%%%%%

%%%%%%%%%%%%%%%%%%%%%%%%%%%%%%%%%%%%%%%%%%%%%%%%%%%%%%%%%%%%%%%%%%%%%%%%%%%%%%%%
%
% 【图】Qwen2.5模型在不同采样策略下，在MATH-500数据集上的性能表现
%
%%%%%%%%%%%%%%%%%%%%%%%%%%%%%%%%%%%%%%%%%%%%%%%%%%%%%%%%%%%%%%%%%%%%%%%%%%%%%%%%

Building on this observation, we report two additional initial accuracy for reference in Fig.~\ref{fig:rlvr_qwen_llama_math500}: (i) \emph{\greedyonly} corresponds to the best performance of the initial model, and (ii) \emph{pass@16}, adopted from \citet{DBLP:DoesRLIncentivizeReasoning}, serves as a plausible performance upper bound for the initial model. Viewed against these baselines, the seeming `RL gains' of \qwenmathSevenB largely reflect adaptation to the template format and merely converge to the \emph{\greedyonly} baseline, indicative of memory recall rather than genuine mathematical generalization.
However, spurious rewards, \emph{e.g.}, random and mv-incorrect, still boost the accuracy of Qwen base and maintain the performance of Qwen instruct, while degrading Llama eventually. This is the point we need to further explore in the following sections.

%%%%%%%%%%%%%%%%%%%%%%%%%%%%%%%%%%%%%%%%%%%%%%%%%%%%%%%%%%%%%%%%%%%%%%%%%%%%%%%%
%
% 【表】三个Base模型在所有数据集上的ROUGEL和EM值
% \input{table/problem_memorization_capability}

\definecolor{Gray}{gray}{0.90}

\begin{table*}[t]
\centering
\small
\setlength{\tabcolsep}{4pt}
\begin{tabular}{llccccccc}
\toprule
\multirow{2}{*}{\textbf{Model}} & \multirow{2}{*}{\textbf{Dataset}} & \multirow{2}{*}{\textbf{Size}} 
& \multicolumn{2}{c}{\textbf{80\%-Problem}} 
& \multicolumn{2}{c}{\textbf{60\%-Problem}} 
& \multicolumn{2}{c}{\textbf{40\%-Problem}} \\
\cmidrule(lr){4-5} \cmidrule(lr){6-7} \cmidrule(lr){8-9}
& & & ROUGE-L & EM & ROUGE-L & EM & ROUGE-L & EM \\
% 表格数据-开始
\midrule
\multirow{6}{*}{\textbf{\qwenmathSevenB}} 
&MATH-500 & 500 & 81.25  & {\textbf{65.80}}  & 78.06  & {\textbf{54.60}}  & 69.01  & \textbf{39.20} \\
&AMC & 83 & 77.38  & {\textbf{55.42}}  & 70.25  & {\textbf{42.17}}  & 75.17  & \textbf{36.14} \\
&AIME2024 & 30 & 74.04  & {\textbf{56.67}}  & 55.31  & 20.00  & 57.72  & 16.67 \\
\cmidrule{2-9}
&AIME2025 & 30 & \cellcolor{Gray}54.71  & \cellcolor{Gray}16.67  & \cellcolor{Gray}34.88  & \cellcolor{Gray}0.00  & \cellcolor{Gray}27.43  & \cellcolor{Gray}0.00 \\
&MinervaMath & 272 & \cellcolor{Gray}36.08  & \cellcolor{Gray}2.94  & \cellcolor{Gray}31.22  & \cellcolor{Gray}0.37  & \cellcolor{Gray}29.35  & \cellcolor{Gray}0.00 \\
&LiveMathBench & 100 & \cellcolor{Gray}42.76  & \cellcolor{Gray}5.00  & \cellcolor{Gray}32.78  & \cellcolor{Gray}0.00  & \cellcolor{Gray}29.97  & \cellcolor{Gray}0.00 \\

\midrule
\multirow{6}{*}{\textbf{\qwenSevenB}} 
&MATH-500 & 500 & 66.42  & {\textbf{40.20}}  & 60.98  & 21.20  & 50.36  & 8.20 \\
&AMC & 83 & 73.24  & {\textbf{49.40}}  & 64.42  & 33.73  & 63.79  & 28.92 \\
&AIME2024 & 30 & 59.80  & \textbf{30.00}  & 48.69  & 13.33  & 44.65  & 10.00 \\
\cmidrule{2-9}
&AIME2025 & 30 & \cellcolor{Gray}54.61  & \cellcolor{Gray}10.00  & \cellcolor{Gray}37.59  & \cellcolor{Gray}0.00  & \cellcolor{Gray}30.30  & \cellcolor{Gray}0.00 \\
&MinervaMath & 272 & \cellcolor{Gray}35.24  & \cellcolor{Gray}2.94  & \cellcolor{Gray}32.35  & \cellcolor{Gray}0.37  & \cellcolor{Gray}27.89  & \cellcolor{Gray}0.00 \\
&LiveMathBench & 100 & \cellcolor{Gray}41.15  & \cellcolor{Gray}4.00  & \cellcolor{Gray}32.74  & \cellcolor{Gray}0.00  & \cellcolor{Gray}27.95  & \cellcolor{Gray}0.00 \\

\midrule
\multirow{6}{*}{\textbf{\llamaEightB}} 
&MATH-500 & 500 & 48.33  & 17.80  & 40.55  & 3.80  & 32.07  & 0.60 \\
&AMC & 83 & 44.54  & 4.82  & 30.62  & 0.00  & 27.10  & 0.00 \\
&AIME2024 & 30 & 50.50  & 13.33  & 30.80  & 0.00  & 26.08  & 0.00 \\
\cmidrule{2-9}
&AIME2025 & 30 & \cellcolor{Gray}47.04  & \cellcolor{Gray}10.00  & \cellcolor{Gray}33.49  & \cellcolor{Gray}0.00  & \cellcolor{Gray}25.20  & \cellcolor{Gray}0.00 \\
&MinervaMath & 272 & \cellcolor{Gray}36.24  & \cellcolor{Gray}2.21  & \cellcolor{Gray}29.52  & \cellcolor{Gray}0.00  & \cellcolor{Gray}27.11  & \cellcolor{Gray}0.00 \\
&LiveMathBench & 100 & \cellcolor{Gray}35.55  & \cellcolor{Gray}5.00  & \cellcolor{Gray}31.93  & \cellcolor{Gray}0.00  & \cellcolor{Gray}26.88  & \cellcolor{Gray}0.00 \\

% 表格数据-结束
\bottomrule
\end{tabular}
\caption{
Accuracy (EM) and ROUGE-L on several datasets (lower scores in \cellcolor{Gray}gray) under different prompt prefix ratios with \textbf{\textit{\greedyonly}} configuration. Suspicious accuracy is highlighted in bold.
}
\label{tab:problem_memorization_greedy}

\end{table*}

%
%%%%%%%%%%%%%%%%%%%%%%%%%%%%%%%%%%%%%%%%%%%%%%%%%%%%%%%%%%%%%%%%%%%%%%%%%%%%%%%%
\subsection{Analysis of Memorization Capability}
\label{sec:memorization-math500}

Considering the Qwen series is trained on massive web-scale corpora, we hypothesize that its divergent RLVR behavior from Llama is because the evaluation set MATH-500 may be inadvertently contaminated in Qwen’s large-scale training data, which is hard to eliminate completely. To verify our hypothesis, we probe memorization on several widely used mathematical-reasoning benchmarks. Concretely, we truncate the original questions at 40\%, 60\%, and 80\% of their lengths and feed these partial questions as prompts into the model, and then evaluate the model’s \textbf{partial-prompt completion rate} by computing ROUGE and EM scores between the generated completion and ground-truth continuations. In addition, we evaluate the model’s \textbf{partial-prompt answer accuracy} by checking if the continuation contains the correct answer, across both partial and full question settings.

The detailed results are presented in Tab.~\ref{tab:problem_memorization_greedy}, revealing strong signs of data contamination in the Qwen2.5 series models when evaluated on commonly used benchmarks, such as MATH-500, AMC, and AIME2024. For instance, when only the first 60\% of the questions are provided, \qwenmathSevenB is able to accurately reconstruct more than half of the remaining problems on MATH-500. Even when just 40\% proportion of the questions are shown, the model still manages to recover 39.2\% of the problems on MATH-500. Similar patterns are observed on AMC and AIME2024. These results indicate that the evaluation benchmarks for Qwen2.5 may suffer from data contamination. Although pre-training on massive web-scale corpora brings strong capacity on mathematical reasoning, \emph{e.g.}, superior performance on recently introduced mathematical tasks like LiveMathBench and AIME2025, those large-scale corpora also include publicly available benchmark problems inevitably, leading to less convincing results of old benchmarks, while removing such instances during large-scale crawling is notoriously difficult.

\begin{figure}[!t]
  \centering
  \includegraphics[width=0.95\linewidth]{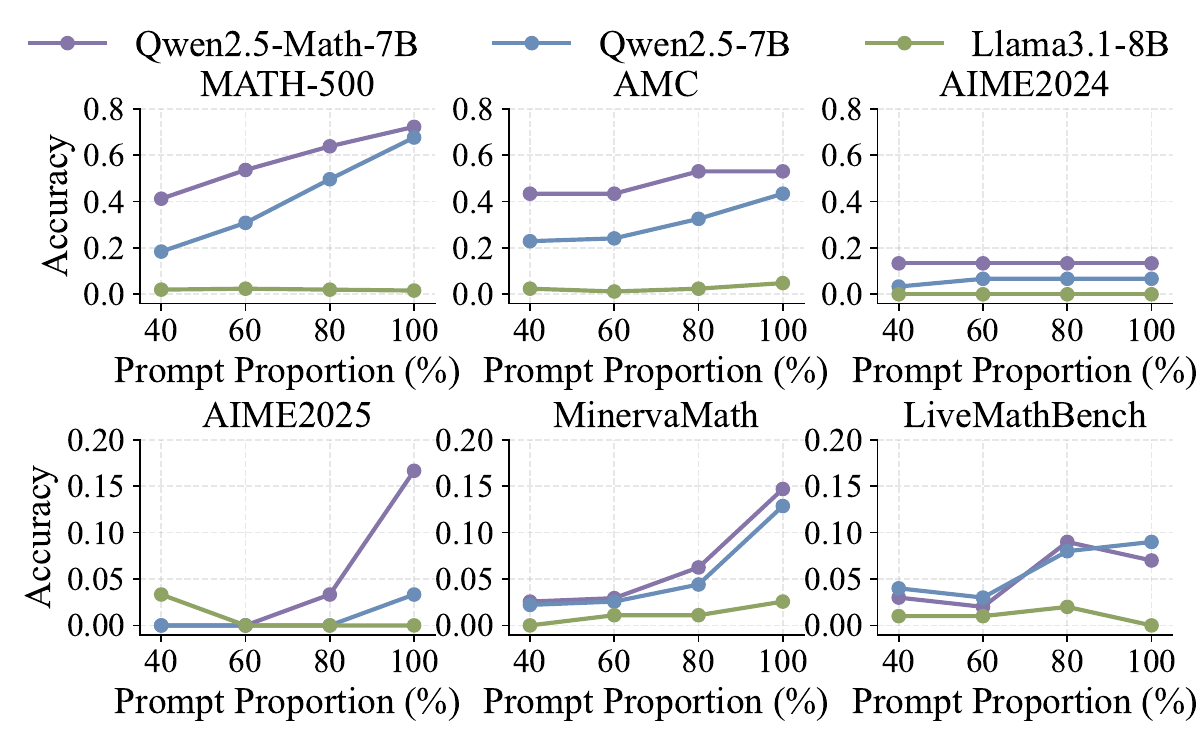}
  \caption{Accuracy (\%) of LLMs on various math datasets under \textbf{\textit{\greedyonly}} configuration. More detailed experimental results can be found in Tab. 5 of Appendix.}
  % 上方有一个附录引用
  % Tab.~\ref{tab:answer_memorization_greedy_only} of Appendix.
  \label{fig:answer_memorization_greedy}
  % \vspace{-0.2in}
\end{figure}

Meanwhile, we summarize the answer accuracy of the model under different ratios of prefix in Fig.~\ref{fig:answer_memorization_greedy}. The Qwen2.5 models achieve remarkably high accuracy on MATH-500 even with partial questions. For instance, with 80\% proportion of the questions, \qwenmathSevenB reaches an accuracy of 63.8\% on MATH-500. Even with only 40\% proportion of the questions, the model still achieves an accuracy of 41.2\%. Because our evaluation matches only the final numeric answer, the model can sometimes output the correct value by accident, which explains the anomalous accuracy of Llama on AIME2025 (solving exactly one problem with a faulty reasoning process).
Besides, we also inspect several questions that Qwen solves correctly, as shown in Fig. 11 to 15 in Appendix. We find that the responses of Qwen contain coherent reasoning chains and even syntactically valid Python code, which, however, is not executed. The emergence of such structured solutions indicates that the training corpora may have included publicly available resources where benchmark problems are accompanied by detailed solutions. Besides, we provide more results for Qwen2.5 and Qwen3 under various sampling configurations in Appendix. The memorization of Qwen2.5 on LiveCodeBench~\citep{arxiv:LiveCodeBench} and Qwen3 series on math is also analyzed in Appendix, showing similar results as Tab.~\ref{tab:problem_memorization_greedy}.
% 上方有附录中的引用：
% as shown in Fig.~\ref{fig:example_qwen2.5_7b_amc} to \ref{fig:example_qwen2.5_math_7b_aime2024}

%%%%%%%%%%%%%%%%%%%%%%%%%%%%%%%%%%%%%%%%%%%%%%%%%%%%%%%%%%%%%%%%%%%%%%%%%%%%%%%%
%
% 【图】三种Base模型在全部六种数据集上，使用不同比例的题目得到的【答案准确率折线图】
%
%%%%%%%%%%%%%%%%%%%%%%%%%%%%%%%%%%%%%%%%%%%%%%%%%%%%%%%%%%%%%%%%%%%%%%%%%%%%%%%%

\subsection{Spurious Rewards on RandomCalculation}

To further support our hypothesis that the anomalous performance surge of the \qwenmathSevenB on the MATH-500 benchmark is primarily caused by data contamination rather than the model’s intrinsic mathematical reasoning ability, we replicate this experiment on a newly constructed dataset that the model has never encountered before. We hypothesize that for math problems free from contamination, the model's reasoning ability still requires properly aligned reward signals to yield meaningful performance improvements. 

\paragraph{Dataset Construction of \randomcalculation.}
To obtain an uncontaminated evaluation benchmark, we employ Algorithm 1 (shown in Appendix) to construct a suite of challenging yet verifiable datasets. These datasets are composed of expressions built from basic numerical elements, including integers from 0 to 100, as well as fractions, squares, and cubes derived from them. Using these components, we randomly generate mathematical expressions that involve between 1 and 20 steps, using the four fundamental arithmetic operations: \textbf{addition}, \textbf{subtraction}, \textbf{multiplication}, and \textbf{division}. To construct the final datasets, we append a standardized problem prefix to each generated expression, resulting in 20 sub-datasets, each containing 1,000 unique problems. We refer to this suite of datasets as \textit{\textbf{\randomcalculation}}. Examples from the datasets can be found in Fig.~\ref{fig:example_random_calculation}.
% 上方有一个附录引用
% we employ Algorithm~\ref{alg:random-calculation}

\begin{figure}
  \centering
  \includegraphics[width=0.4\textwidth]{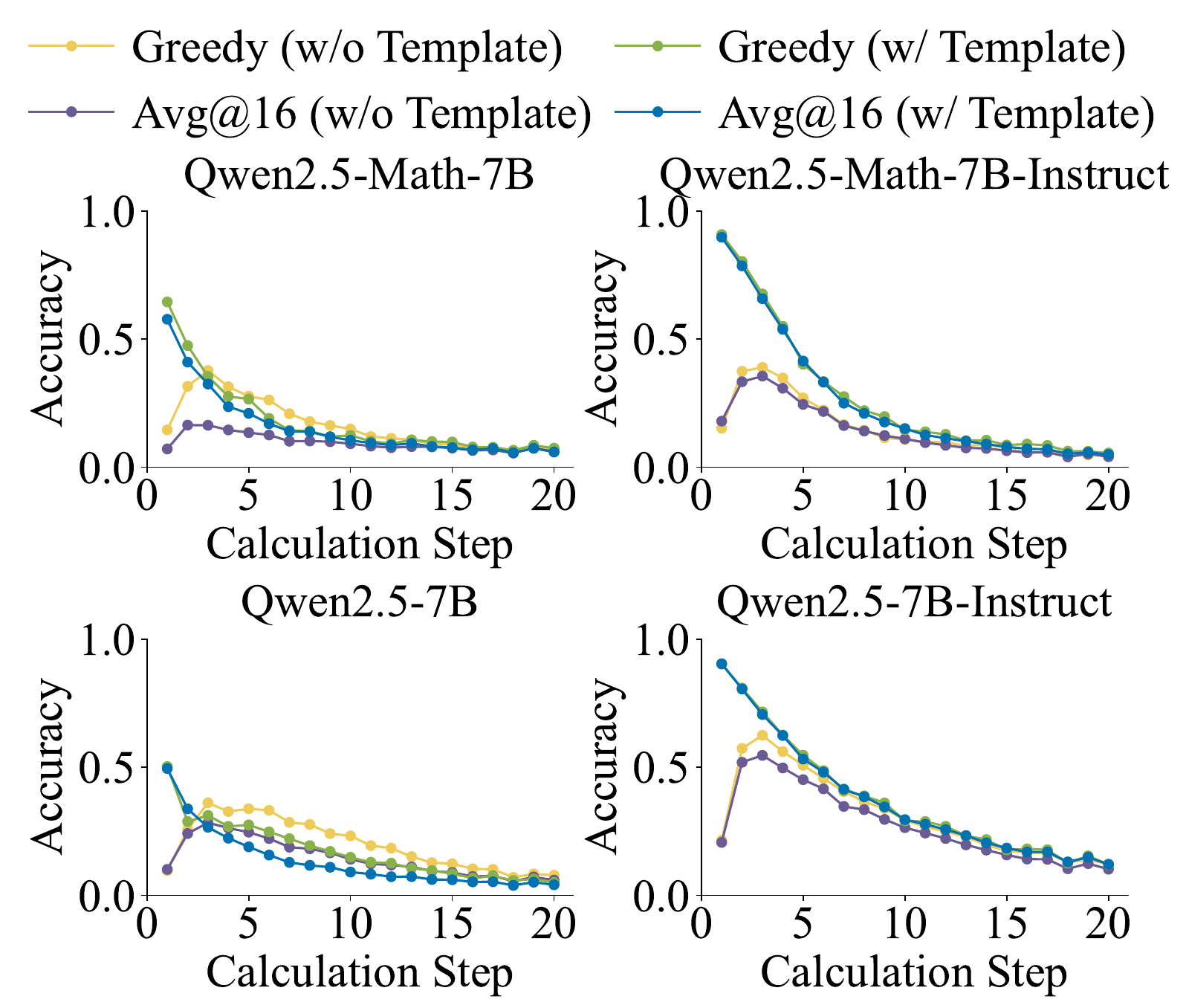}
  \caption{Performance of the Qwen2.5 series models on the \textit{\randomcalculation} datasets under different configurations.}
  % The configuration parameters are listed in Table~\ref{tab:gen-configs}.}
  \label{fig:random_calc_datasets}
  % \vspace{-0.2in}
\end{figure}

\paragraph{Zero-shot Performance on {\randomcalculation}.}
We first test the zero-shot performance of the Qwen2.5 series on the \textit{{\randomcalculation}} dataset in Fig.~\ref{fig:random_calc_datasets}. We find that when using chat templates, the models' performance degrades gradually as the number of computation steps increases, leaving ample room for improvement in multi-step calculation problems. When chat templates are removed, the reasoning performance peaks on problems of three computation steps, and then gradually declines.

%%%%%%%%%%%%%%%%%%%%%%%%%%%%%%%%%%%%%%%%%%%%%%%%%%%%%%%%%%%%%%%%%%%%%%%%%%%%%%%%
%
% 【图】Qwen2.5系列模型在RandomCalculation数据集上的准确率走势
%
%%%%%%%%%%%%%%%%%%%%%%%%%%%%%%%%%%%%%%%%%%%%%%%%%%%%%%%%%%%%%%%%%%%%%%%%%%%%%%%%
\paragraph{Correct Reward Function for \textit{RandomCalculation}.}
\label{sec:reward-function}
The ground-truth answers to our randomly generated arithmetic problems often contain high-precision decimals. When using the standard RLVR framework, which only provides binary feedback, the model rarely receives positive reinforcement, making training unstable and prone to divergence. To overcome this, we design a continuous reward function that ranges from 0 to 1 and penalizes both absolute and relative errors between the model’s prediction and the reference answer. This richer feedback helps stabilize reinforcement learning. Let \( a \) be the model output, \( b \) be the reference answer, and \( \epsilon = 10^{-6} \) be a small constant for numerical stability. The reward \( r \) is computed as:

\begin{equation}
r = 1 
- \underbrace{0.5 \cdot \min\left( |a - b|,\ 1 \right)}_{\text{absolute distance}} 
- \underbrace{0.5 \cdot \min\left( \frac{|a - b|}{|b| + \epsilon},\ 1 \right)}_{\text{relative distance}}
\label{eq:reward}
\end{equation}

\begin{figure}
  \centering
  \includegraphics[width=\linewidth]{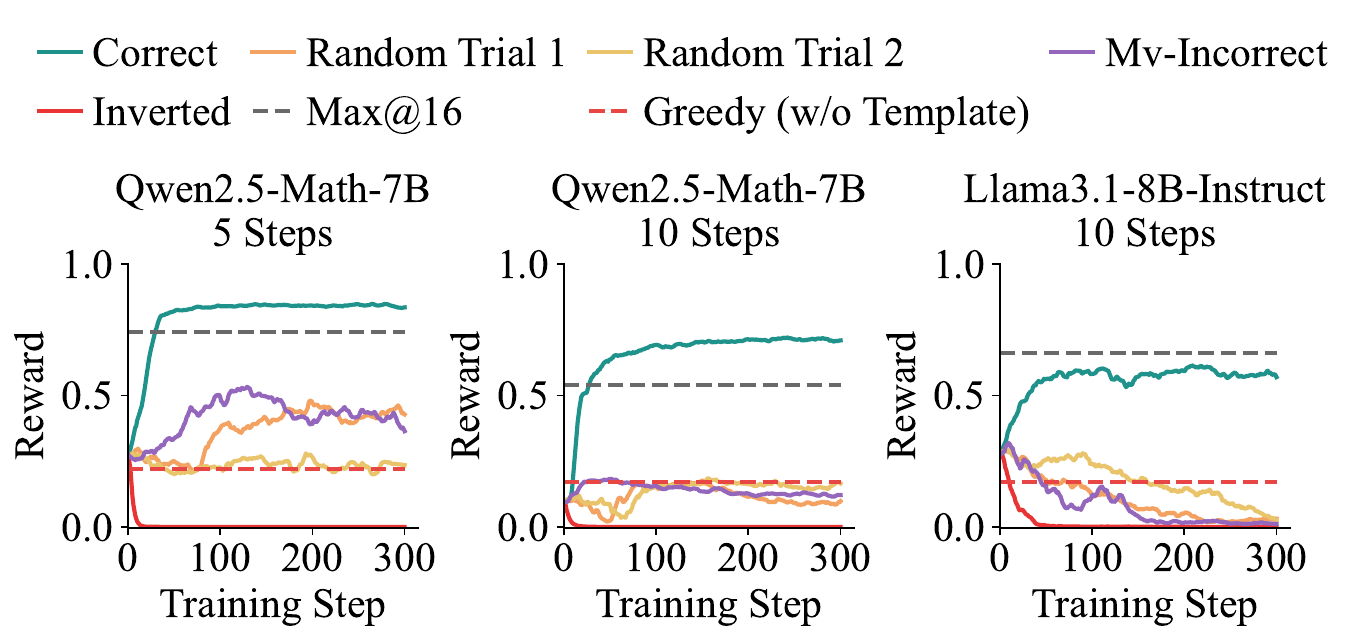}
  \caption{Reward of {\qwenmathSevenB} and {\llamaEightBInstruct} on \textit{\randomcalculation}. Results are presented for datasets with 5-step and 10-step calculations.}
  \label{fig:qwen_random_calc_and_rlvr}
\end{figure}

\paragraph{RLVR on {\randomcalculation}.}
\label{sec:rlvr-random-calculation}
We also perform RLVR training on \qwenmathSevenB using the \textit{\randomcalculation} datasets.
% , aiming to guide the model toward producing answers that are as close as possible to the correct ones. 
Specifically, experiments are conducted on two sub-datasets comprising 5-step and 10-step calculation problems. Each dataset contains 1,000 problems, with 700 used for training and the remaining 300 reserved for validation. As shown in Fig.~\ref{fig:qwen_random_calc_and_rlvr}, the performance improves steadily throughout training under correct rewards. However, training becomes unstable and inconsistent with random or incorrect rewards. Under inverted rewards, the performance collapses rapidly. These findings suggest that for problems not leaked during pretraining, only correct reward signals can effectively guide the model toward improved performance. For comparison, we also evaluate {\llamaEightBInstruct} and observe similar findings. 

%%%%%%%%%%%%%%%%%%%%%%%%%%%%%%%%%%%%%%%%%%%%%%%%%%%%%%%%%%%%%%%%%%%%%%%%%%%%%%%%
%
% 【图】Qwen2.5-Math-7B在RandomCalculation数据集上使用RLVR算法的训练表现。这里展示了5和10两个计算步骤数据集的结果。
%
%%%%%%%%%%%%%%%%%%%%%%%%%%%%%%%%%%%%%%%%%%%%%%%%%%%%%%%%%%%%%%%%%%%%%%%%%%%%%%%%

\paragraph{Qwen v.s. Llama on Clean Benchmark.} Considering obtaining a reward of 1 on a \randomcalculation instance is virtually impossible, we report \textit{Max@16}, the highest reward among 16 samples of initial model, in Fig.~\ref{fig:qwen_random_calc_and_rlvr}. We observe that on \randomcalculation datasets, \qwenmathSevenB can surpass Max@16 when provided with correct reward signals. This finding indicates that reward-aligned RLVR effectively transfers the high-accuracy single-step arithmetic skills (as shown in Fig.~\ref{fig:random_calc_datasets}) to more complex multi-step calculations, as illustrated by one reasoning trace in Fig. 16 of Appendix. In contrast, under the incorrect and random reward configurations, the Qwen model either maintains its base performance or exhibits only marginal and unstable improvements, due to the learning of format as explained in \citet{Spurious-Rewards}. This improvements gradually disappear when calculation steps increased from 5 to 10. This further highlights the critical role of correct and well-aligned reward signals in enhancing model performance on uncontaminated and difficult datasets. Notably, \llamaEightBInstruct fails to surpass the Max@16 even when trained with correct reward signals, and its accuracy falls below the greedy-decoding baseline when exposed to spurious signals. This discrepancy indirectly suggests that Qwen2.5 exhibits stronger mathematical capabilities than Llama3.1 before mid-training. However, such inherent strength is not the root cause of its performance boosting under spurious reward on contaminated datasets.
% 上方有一个附录引用
% one reasoning trace in Fig.~\ref{fig:example_rlvr_qwen2.5_math_7b_5_step_correct}

\subsection{More Evidence for Memorization}
\label{subsec:token_memorisation}
Here, we provide more detailed analyses of Qwen's sudden performance gains on MATH-500 under random reward. Let $
\mathcal{J}_{\text{CLIP}}
  =\mathbb{E}_{\hat{A}_{i,t}}\!\Bigl[\min \left(r_{i,t} \hat{A}_{i,t}, \text{clip} \left( r_{i,t}, 1 - \epsilon, 1 + \epsilon \right)\hat{A}_{i,t} \right)\Bigr],
$ where $\hat{A}_{i,t}$ is a random variable under the setup of random reward.
Referring to Appendix B of ~\citet{Spurious-Rewards}, the gradient of the clipped policy has the following format:
\begin{equation}
\nabla_\theta J_{\text{CLIP}}=
\nabla_\theta r_{i,t}\cdot G(r_{i,t}),
\end{equation}
\begin{equation}
G(r_{i,t})=
\begin{cases}
  \mu, & r_{i,t} < 1-\epsilon,\\
  0, & 1-\epsilon \le r_{i,t} \le 1+\epsilon,\\
 -\mu, & r_{i,t} > 1+\epsilon,\\
\end{cases}
\label{eq:g_rt_piecewise}
\end{equation}
where $\mu>0$ is a positive coefficient, $r_{i,t}=\frac{\pi_\theta(o_{i,t} | q, o_{i,<t})}{\pi_{\theta_{\text{old}}}(o_{i,t} | q, o_{i,<t})}$.
% where the indicator $\mathbf{1}_{\,|r_t-1|\le\epsilon}$ nullifies the gradient whenever
% $r_t$ leaves the clipping interval.

\paragraph{Memory Retrieval due to Exploitation Bias.}  
Assume a high-probability token with $\pi_{\text{old}}=0.85$ and $\epsilon=0.20$, the upper clipping boundary is $1.02$ for $\pi_\theta$, which exceeds the probability ceiling of $1.0$ and therefore is never reached. Consequently, the gradient is non-negative for this token, leading to a net positive gradient bias on the policy model,
In general, for high-probability token, we have
$
\nabla_\theta J_{\text{CLIP}}(\theta)
  \propto
  \nabla_\theta r_{i,t}$, due to $G(r_{i,t})\ge 0$ hold almost surely. So high‐probability tokens continue to be up-weighted without penalty. For \textsc{MATH‐500}, correct answers typically have a high probability due to data contamination in the initial model (results shown in Fig.9 of Appendix), except for the low-probability answer format. Therefore, GRPO with random reward can retrieve these answers after learning format and leads to sharp accuracy jump in Fig~\ref{fig:rlvr_qwen_llama_math500}.
% 上方有一个附录引用
% results shown in Fig.\ref{fig:token_prob_change} of Appendix

On the other hand, assume another token with pre-update likelihood $\pi_{\text{old}}=0.5$ (which is a typical value for our 10-step \randomcalculation as shown in Fig. 9 of Appendix).
% 上方有一个附录引用
% as shown in Fig.~\ref{fig:token_prob_change} of Appendix
The corresponding clipping boundary is $[0.4, 0.6]$ for $\pi_\theta$. 
Gradient update with random reward perturbs $\pi_{\theta}$ around this narrow band, so that $G(r_{i,t})\approx 0$ in most cases. Consequently,
$
\nabla_\theta J_{\text{CLIP}}(\theta)\approx\mathbf{0}.
$ Therefore, no meaningful performance improvement observed in 10-step \randomcalculation with random reward in Fig.~\ref{fig:qwen_random_calc_and_rlvr}.
Overall, clipped objective introduces systematic \emph{exploitation bias} for high-probability tokens,
% , \emph{i.e.}, under random reward for training freely, a high-probability `exploitation' token is less constrained by upper clipping and can achieve a larger probability
whereas mid-probability tokens are less optimized.
% Empirically, responses of 10-step \randomcalculation contain many digits with probability of $0.5$ in Appendix Fig.\ref{fig:token_prob_change}, which display no consistent amplification or suppression of these tokens after RL with random reward, corroborating the limited or unstable gains shown in Fig.~\ref{fig:qwen_random_calc_and_rlvr}.

\begin{figure}
  \centering
  \includegraphics[width=\linewidth]{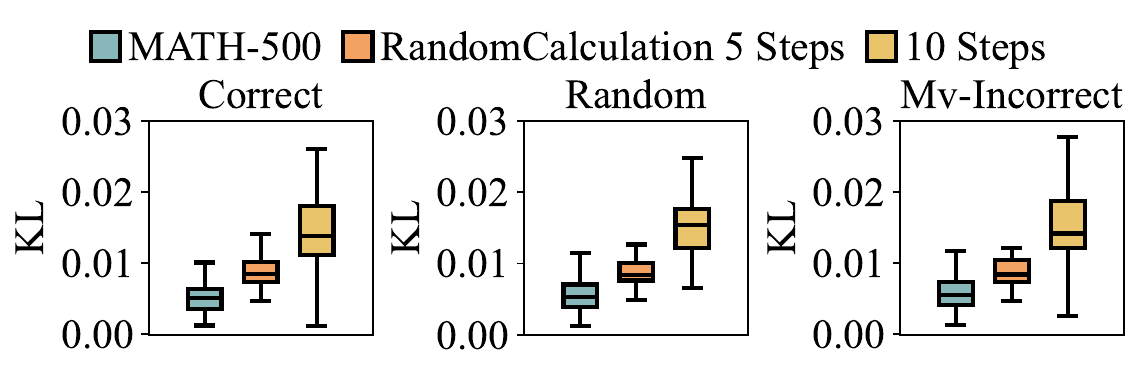}
  \caption{KL distance of model outputs before and after RL.}
  \label{fig:kl_divergence}
  % \vspace{-0.2in}
\end{figure}

\paragraph{Response Similarity Before and After RL.}
We further compare the responses of the model before and after RL, with \textsc{Rouge-L} and KL distance\footnote{For each generated answer we compute $\mathrm{KL}\!\left(\mathrm{P}_{\text{Base}}\;\Vert\;\mathrm{P}_{\text{FT}}\right)$ over the full vocabulary, where $\mathrm{P}_{\text{Base}}$ and $\mathrm{P}_{\text{FT}}$ denote probability distributions produced by the fine-tuned and base models, respectively.} as the similarity score. As shown in Tab.~\ref{tab:rouge_results} and Fig.~\ref{fig:kl_divergence}, the similarity in MATH-500 is substantially higher than in \randomcalculation, further implying that MATH-500 suffers from data contamination. Additionally, spurious rewards achieve even higher \textsc{Rouge-L} than correct reward after RL. 
% Besides, we also compute the token-level KL distance between the RL model and initial model.  
% As shown in , the overall KL on MATH-500 is considerably lower than \randomcalculation under various rewards.
% This alignment lends further support to our hypothesis: Qwen2.5 inadvertently memorized substantial portions of MATH-500 during pre-training, and RLVR training predominantly strengthens chain-of-thought retrieval and answer extraction from memory rather than fostering novel reasoning skills.
Therefore, performance surge under spurious rewards arises because GRPO inadvertently triggers Qwen to retrieve memorized answers, rather than stimulating Qwen’s existing reasoning patterns like codes as explained in \citet{Spurious-Rewards}. This is due to the exploitation bias of GRPO. However, RL with correct reward can still stimulate the model to find new reasoning paths, as verified with a smaller \textsc{Rouge-L} of response.

\begin{table}
\centering
\small
\setlength{\tabcolsep}{4pt}  % 调小列间距
%
% 不同的生成配置使用的vLLM采样参数
%
\begin{tabular}{lcccccc}
\toprule
\textbf{Dataset} & \textbf{Reward Signal} & \textbf{ROUGE-L} \\
\midrule
\multirow{3}{*}{MATH-500}
    & Correct & 0.555 \\
    & Random & \textbf{0.601} \\
    & Mv-Incorrect & 0.563 \\
\midrule
\multirow{3}{*}{\randomcalculation 5 Steps}
    & Correct & 0.225 \\
    & Random & 0.247 \\
    & Mv-Incorrect & \textbf{0.251} \\
\midrule
\multirow{3}{*}{\randomcalculation 10 Steps}     
    & Correct & 0.193 \\
    & Random & 0.251 \\
    & Mv-Incorrect & \textbf{0.279} \\
\bottomrule
\end{tabular}
\caption{Similarity of model outputs before and after RL.}
\label{tab:rouge_results}
% \vspace{-0.2in}
\end{table}

% \paragraph{Token–Level KL Distance Before and After RL.}
% To gain a finer-grained analysis, we randomly sample 100 instances from MATH-500 and \randomcalculation and , further implying that memorization is the main source of performance improvement on MATH-500.
% Under the random reward condition, the median KL value is below~$0.01$, indicating that the performance improvement on MATH-500 is attributable primarily to template adaptation and answer-pattern memorization exacerbated by the known \emph{CLIP} bias, rather than to genuine mathematical reasoning.

%%%%%%%%%%%%%%%%%%%%%%%%%%%%%%%%%%%%%%%%%%%%%%%%%%%%%%%%%%%%%%%%%%%%%%%%%%%%%%%%
%
% 【图】Qwen2.5-Math-7B在RandomCalculation数据集上使用RLVR算法的训练表现。这里展示了5和10两个计算步骤数据集的结果。
%
%%%%%%%%%%%%%%%%%%%%%%%%%%%%%%%%%%%%%%%%%%%%%%%%%%%%%%%%%%%%%%%%%%%%%%%%%%%%%%%%

% To probe this further, we inspected the probabilities that the base model assigns to the numerical tokens appearing in the final answers produced by the fine-tuned models. 
% For MATH-500, these numerical tokens already receive probabilities approaching~1 in the base model.  
% Conversely, in \randomcalculation, the numerical tokens produced by the fine-tuned models are assigned substantially lower probabilities by the base model, reinforcing the interpretation that the improvements on MATH-500 stem from strengthened probability again.

% \section{Discussion and Limitation}

% 最近一段时间所提出的用于提升数学推理能力的RL算法非常多，我们在短期内并不能对所有算法进行验证。
% 

\section{Conclusion}

In this work, we investigate the unexpected performance improvements of Qwen on mathematical reasoning with spurious rewards. Our analysis reveals that these gains were primarily due to data contamination rather than Qwen's inherent mathematical capabilities. By auditing the MATH-500 dataset and introducing a clean benchmark, we demonstrate that Qwen's successes with spurious reward were driven by memorization of benchmark problems rather than genuine reasoning skills. Additionally, we show that only correctly aligned rewards lead to consistent performance improvements, while spurious rewards fail to provide meaningful benefits. These findings underscore the importance of using uncontaminated benchmarks in evaluating RL-based methods and call for caution when interpreting results from datasets that may suffer from data leakage. Our work highlights the need for rigorous evaluation protocols in future research to ensure that performance gains reflect true advancements in ability, rather than data contamination.

\section*{Acknowledgments}
The authors wish to thank the anonymous reviewers for their helpful comments. This work was partially funded by National Natural Science Foundation of China (No.62476061, 62376061, 62206057), Shanghai Rising-Star Program (23QA1400200), Natural Science Foundation of Shanghai (23ZR1403500). The computations were partially performed using Ascend AI Accelerators. The authors would like to thank Ascend Cloud Ecological Development Project for the support of Ascend 910 processors. Qin Liu is supported by the Amazon Nova Trusted AI Prize.

\bibliography{aaai2026}

% If so, you can uncomment the following line and adjust the path to include it.
% Camera-Ready版本不需要加Checklist

%%%%%%%%%%%%%%%%%%%%%%%%%
% split pdf here
\clearpage
\newpage
\appendix
\section{Discussion and Limitation}
Due to limited computational resources, our experiments were restricted to a subset of commonly used Qwen2.5 and Qwen3 series. 
Besides, given the rapid development of various RL algorithms, it is infeasible to conduct a comprehensive evaluation of all these methods in the short term. In future work, our efforts will focus on comprehensive evaluation on more diverse benchmarks, reinforcement learning methods, and model families. 
% In parallel, we will carry out an in-depth theoretical investigation of spurious rewards, with the goal of shedding light on their root causes and advancing our understanding of the underlying mechanisms.

\section{Details of \randomcalculation Construction}
We provide the specific algorithm for the construction of our clean \randomcalculation benchmark here:
%%%%%%%%%%%%%%%%%%%%%%%%%%%%%%%%%%%%%%%%%%%%%%%%%%%%%%%%%%%%%%%%%%%%%%%%%%%%%%%%
%
% 【算法】四则运算数据集构造过程算法
%
\begin{algorithm}[!ht]
\caption{Construction of \textit{{\randomcalculation}} Dataset}
\label{alg:random-calculation}
\begin{algorithmic}
\REQUIRE Maximum computation steps: $\textbf{N} = 20$
\STATE Initialize dataset $S_0$ with basic mathematical expressions
\STATE Initialize dataset list: $\textbf{DL} \leftarrow \{S_0\}$
\STATE Define operator set: $\textbf{OPSET} \leftarrow \{+, -, \times, \div\}$
\FOR{$i = 1$ to $N$}
    \STATE $D_i \leftarrow \emptyset$
    \FOR{$j = 0$ to $\lceil i/2 \rceil$}
        \STATE Randomly select $\textbf{Left} \in \textbf{DL}[j]$
        \STATE Randomly select $\textbf{Right} \in \textbf{DL}[i - 1 - j]$
        \STATE Randomly select $op \in \textbf{\text{OPSET}}$
        \STATE Randomly swap $\textbf{Left}$ and $\textbf{Right}$
        \STATE $\textbf{expr} \leftarrow \textbf{Left} \; op \; \textbf{Right}$
        \STATE Add $\textbf{expr}$ to $D_i$
    \ENDFOR
    \STATE Append $D_i$ to $\textbf{DL}$
\ENDFOR
\STATE Save $\textbf{DL}$ as the \textit{{\randomcalculation}} dataset
\end{algorithmic}
\end{algorithm}

%
%%%%%%%%%%%%%%%%%%%%%%%%%%%%%%%%%%%%%%%%%%%%%%%%%%%%%%%%%%%%%%%%%%%%%%%%%%%%%%%%

% 数字答案token概率量化分析
% 图xxx展示了在MATH-500数据集和RandomCalculation 10步骤数据集的100条样本上，与答案相关的数字token在RL前后的概率变化。可以看到MATH-500在RL前后表现出较高的概率，这说明模型对答案有着较高的记忆。而对于无污染的RandomCalculation 10步骤数据集，则表现比较分散，表明随机奖励信号不生效。
\section{Quantitative Analysis of Answer-Related Numeric Token Probabilities}
\label{sec:num_token_prob_change}
\begin{figure*}
  \centering
  \includegraphics[width=\linewidth]{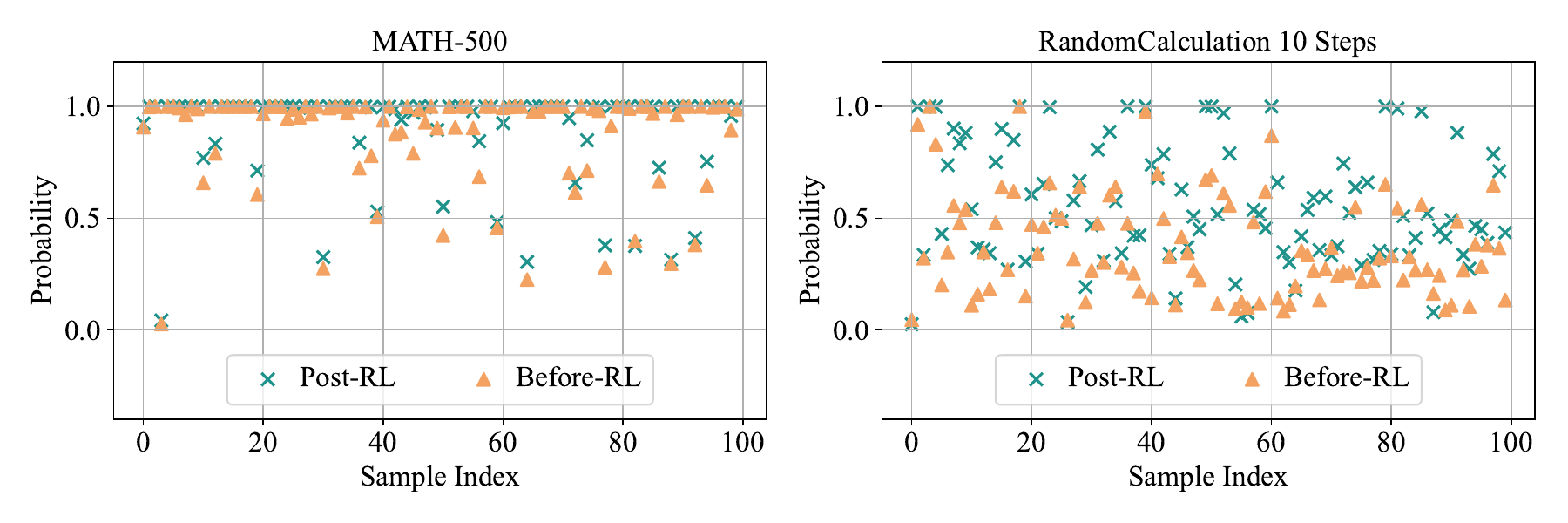}
  % 在MATH-500和RandomCalculation十步骤数据集上，答案数字的Token在RL前后的概率对比
  \caption{Token-level comparison of the probabilities assigned to answer-relevant numeric tokens before and after reinforcement learning (RL) on the MATH-500 and \textit{RandomCalculation} (10-step) benchmarks under a random-reward setting. For each benchmark, 100 problems are randomly selected; the post-RL model generates an answer for every problem, and the probabilities that both the pre- and post-RL models assign to each numeric token appearing in these answers are subsequently evaluated.}
  \label{fig:token_prob_change}
\end{figure*}

Fig. \ref{fig:token_prob_change} illustrates the change in probabilities of numeric tokens associated with final answers before and after reinforcement learning (RL), evaluated on 100 samples from the MATH-500 and 10-step RandomCalculation datasets using random reward. The MATH-500 results exhibit consistently high probabilities both before and after RL, suggesting that the model retains a strong memory of the answers. In contrast, the results on the clean 10-step RandomCalculation dataset are more dispersed, showing that random reward signals are ineffective.

\section{Quantitative Results on Partial-Prompt Answer Accuracy} 
\label{sec:appendix-ppaa-Qwen2.5}
Tab.~\ref{tab:math-500-comparison} presents an expanded comparison of partial-prompt answer accuracy between Qwen-2.5 and Llama-3.1 on the \textsc{MATH-500} benchmark.  To complement these findings, Tab.~\ref{tab:answer_memorization_greedy_only}–\ref{tab:answer_memorization_template_sampling} reports analogous results across several additional mathematics datasets, evaluated under four generation configurations: Greedy decoding with and without templates, and Avg@16 decoding with and without templates. 

\section{Quantitative Results for Qwen3} 
\label{sec:appendix-Qwen3}
Qwen3 is the closest open-source model within the Qwen family, and we subject it to the same memorization diagnostics. Table~\ref{tab:qwen3_problem_memorization_greedy} reports its partial-prompt completion rate, whereas Tables~\ref{tab:qwen3_answer_memorization_greedy_only}–\ref{tab:qwen3_answer_memorization_template_sampling} present the corresponding partial-prompt answer accuracy under all generation settings. The results mirror those of Qwen2.5: despite the model’s increased capacity, it still exhibits pronounced evidence of data contamination, likely attributable to pre-training on large-scale web corpora.

\section{RLVR on LiveMathBench}
\label{sec:appendix-LiveMathBench}
\begin{figure*}
  \centering
  \includegraphics[width=0.35\linewidth]{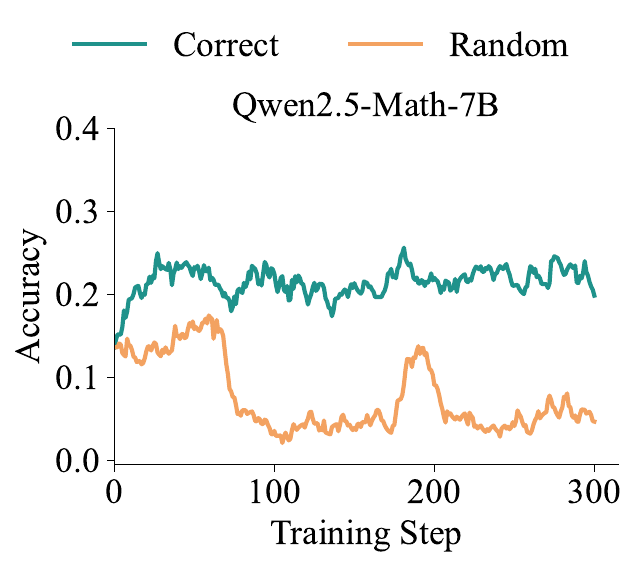}
  % 使用LiveMathBench（v202412+v202505）对Qwen2.5-Math-7B进行RLVR训练
  \caption{
    {RLVR training is performed on Qwen2.5-Math-7B with the LiveMathBench (v202412 + v202505) dataset under correct and random reward signals.}
  }
  \label{fig:rlvr_qwen_livemathbench}
\end{figure*}

In the early exploratory phase, we conduct RLVR training on \qwenmathSevenB with the clean LiveMathBench dataset. The experimental results are shown in Fig. \ref{fig:rlvr_qwen_livemathbench}. Under correct reward signal, the model achieves a limited improvement, which is mainly due to the relatively small amount of training data. In contrast, when using random reward signal, the model fails to obtain stable performance gains and ultimately exhibits a declining trend.

\section{Other Reasoning Domains}
\label{sec:appendix-code-domain}
Beyond the mathematics domain, we also conduct preliminary memory tests on LiveCodeBench~\citep{arxiv:LiveCodeBench}, a commonly used code evaluation benchmark. The results are shown in Tab. \ref{tab:qwen_code_memorization_greedy}. When 80\% proportion of problems provided as prompts, Qwen2.5-Math-7B is able to accurately reproduce 56.59\% of the problems, whereas Llama3.1-8B only reproduces 4.40\%. We also observe that the outputs of Qwen2.5-Math-7B frequently included complete test cases, as illustrated in Fig. \ref{fig:example_qwen2.5_math_7b_livecodebench}.

\section{Illustrative Examples of Model Output}
\label{sec:appendix-examples}
Figures \ref{fig:example_qwen2.5_7b_amc} and \ref{fig:example_qwen2.5_7b_aime2024} present concrete instances of memorization by Qwen2.5-7B on the AMC and AIME2024 benchmarks, respectively.
Figures \ref{fig:example_qwen2.5_math_7b_math_500}–\ref{fig:example_qwen2.5_math_7b_aime2024} provide analogous examples for Qwen2.5-Math-7B on MATH-500, AMC, and AIME2024.
Figure \ref{fig:example_rlvr_qwen2.5_math_7b_5_step_correct} depicts a representative response produced by the RLVR-fine-tuned Qwen2.5-Math-7B on the 5-step \randomcalculation benchmark.

\section{Hardware and Software Requirements}
All experiments in this work were conducted on NVIDIA A800 80G GPUs, with Ubuntu 20.04.6 LTS as the operating system and CUDA driver version 12.4. Reinforcement learning (RL) training is performed using 8 A800 GPUs and 512 GB of RAM.

\section{RLVR Training Details}
We follow the standard TTRL experimental configuration. Specifically, during training, the learning rate is set to 5e-7 and the temperature is set to 1.0. For each prompt, we generate 16 samples. The training batch size is set to 128. To improve efficiency and reduce memory usage, we enable FlashAttention to accelerate attention computation in the Transformer model.

%%%%%%%%%%%%%%%%%%%%%%%%%%%%%%%%%%%%%%%%%%%%%%%%%%%%%%%%%%%%%%%%%%%%%%%%%%%%%%%%%%%%%
%%%%%%%%%%%%%%%%%%%%%%%%%%%%%%%%%%%%%%%%%%%%%%%%%%%%%%%%%%%%%%%%%%%%%%%%%%%%%%%%%%%%%
% 
% 以下内容是附录，暂时放在这里
%
%%%%%%%%%%%%%%%%%%%%%%%%%%%%%%%%%%%%%%%%%%%%%%%%%%%%%%%%%%%%%%%%%%%%%%%%%%%%%%%%%%%%%
%%%%%%%%%%%%%%%%%%%%%%%%%%%%%%%%%%%%%%%%%%%%%%%%%%%%%%%%%%%%%%%%%%%%%%%%%%%%%%%%%%%%%

%%%%%%%%%%%%%%%%%%%%%%%%%%%%%%%%%%%%%%%%%%%%%%%%%%%%%%%%%%%%%%%%%%%%%%%%%%%%%%%%
%
% MATH-500和十步骤运算，数字Token的变化
%
%%%%%%%%%%%%%%%%%%%%%%%%%%%%%%%%%%%%%%%%%%%%%%%%%%%%%%%%%%%%%%%%%%%%%%%%%%%%%%%%

%%%%%%%%%%%%%%%%%%%%%%%%%%%%%%%%%%%%%%%%%%%%%%%%%%%%%%%%%%%%%%%%%%%%%%%%%%%%%%%%
%
% Qwen2.5-Math-7B在LiveMathBench上的训练结果
%
%%%%%%%%%%%%%%%%%%%%%%%%%%%%%%%%%%%%%%%%%%%%%%%%%%%%%%%%%%%%%%%%%%%%%%%%%%%%%%%%

%
% Qwen2.5模型在不同采样策略下，在MATH-500数据集上的性能表现
%
%
% 【表格】：MATH-500在全模型全配置下的准确率
%
\definecolor{Gray}{gray}{0.90} 

\begin{table*}[!ht]
\centering
\small
\setlength{\tabcolsep}{6pt}
\begin{tabular}{llcccc}
\toprule
\textbf{Model} & \textbf{Configuration} & \textbf{100\%} & \textbf{80\%} & \textbf{60\%} & \textbf{40\%} \\
\midrule
\multirow{4}{*}{\textbf{\qwenmathSevenB}}
  & \greedyonly          & 72.20  & 63.80  & 53.60  & 41.20  \\
  & \sampleonly        & 68.53 & 57.25 & 45.51 & 31.03 \\
  & \cellcolor{Gray}\templategreedy   & \cellcolor{Gray}50.60  & \cellcolor{Gray}29.20  & \cellcolor{Gray}20.20  & \cellcolor{Gray}10.00  \\
  & \cellcolor{Gray}\templatesample & \cellcolor{Gray}48.84 & \cellcolor{Gray}28.01 & \cellcolor{Gray}18.99 & \cellcolor{Gray}10.36 \\
\midrule
\multirow{4}{*}{\textbf{\qwenmathSevenBInstruct}}
  & \greedyonly          & 83.60  & 53.80  & 31.80  & 16.00  \\
  & \sampleonly        & 81.76 & 50.80 & 31.54 & 15.43 \\
  & \cellcolor{Gray}\templategreedy   &  \cellcolor{Gray}82.20  & \cellcolor{Gray}3.80  & \cellcolor{Gray}24.40  & \cellcolor{Gray}13.20  \\
  &  \cellcolor{Gray}\templatesample &  \cellcolor{Gray}82.01 &  \cellcolor{Gray}43.66 &  \cellcolor{Gray}25.14 &  \cellcolor{Gray}12.78 \\
\midrule
\multirow{4}{*}{\textbf{\qwenSevenB}}
  & \greedyonly          & 67.60  & 49.60  & 30.80  & 18.40  \\
  & \sampleonly        & 66.20 & 46.04 & 30.01 & 16.10 \\
  & \cellcolor{Gray}\templategreedy   & \cellcolor{Gray}40.00  & \cellcolor{Gray}22.20  & \cellcolor{Gray}13.40  & \cellcolor{Gray}6.80  \\
  & \cellcolor{Gray}\templatesample & \cellcolor{Gray}38.15 & \cellcolor{Gray}21.75 & \cellcolor{Gray}11.89 &  \cellcolor{Gray}5.25 \\
\midrule
\multirow{4}{*}{\textbf{\qwenSevenBInstruct}}
  & \greedyonly          & 72.80  & 50.00  & 31.20  & 15.20  \\
  & \sampleonly        & 74.90 & 50.18 & 30.88 & 15.79 \\
  & \cellcolor{Gray}\templategreedy   & \cellcolor{Gray}72.20  & \cellcolor{Gray}36.00  & \cellcolor{Gray}20.80  & \cellcolor{Gray}10.00  \\
  & \cellcolor{Gray}\templatesample & \cellcolor{Gray}73.69 & \cellcolor{Gray}37.16 & \cellcolor{Gray}20.67 &  \cellcolor{Gray}8.75 \\
\midrule
\multirow{4}{*}{\textbf{\llamaEightB}}
  & \greedyonly          &  1.60  & 2.00  & 2.40  & 2.00  \\
  & \sampleonly        &  3.12 &  2.81 &  2.80 &  2.30 \\
  & \cellcolor{Gray}\templategreedy   &   \cellcolor{Gray}--  &   \cellcolor{Gray}--  &   \cellcolor{Gray}--  &   \cellcolor{Gray}--  \\
  & \cellcolor{Gray}\templatesample &   \cellcolor{Gray}--  &   \cellcolor{Gray}--  &   \cellcolor{Gray}--  &   \cellcolor{Gray}--  \\
\midrule
\multirow{4}{*}{\textbf{\llamaEightBInstruct}}
  & \greedyonly          & 43.60  & 24.80  & 15.00 & 7.40 \\
  & \sampleonly        & 39.61 & 24.76 & 14.21 &  6.84 \\
  & \cellcolor{Gray}\templategreedy   & \cellcolor{Gray}38.80  & \cellcolor{Gray}17.20  & \cellcolor{Gray}10.60 & \cellcolor{Gray}5.00 \\
  & \cellcolor{Gray}\templatesample & \cellcolor{Gray}38.24 & \cellcolor{Gray}18.46 &  \cellcolor{Gray}9.95 &  \cellcolor{Gray}4.25 \\
\bottomrule
\end{tabular}
% 在不同的生成配置条件下，以不同比例的题目作为Prompt，Qwen和llama模型在MATH-500数据集上的准确率(%)
\caption{Accuracy (\%) of \textbf{Qwen} and \textbf{Llama} models on the \textbf{MATH-500} dataset under different generation configurations, using varying proportions of questions as prompts. The configuration parameters can be found in Table \ref{tab:gen-configs}. Note that \textit{{\llamaEightB}} lacks an official chat template, so template-dependent experiments are omitted.}
\label{tab:math-500-comparison}
\end{table*}

%%%%%%%%%%%%%%%%%%%%%%%%%%%%%%%%%%%%%%%%%%%%%%%%%%%%%%%%%%%%%%%%%%%%%%%%%%%%%%%%
%
% 【表】全部模型全部Benchmark在所有配置下，不同比例题目作为prompt的准确率（表格形式）
%
% 【表格】：六种模型在贪婪模式下的答案背诵能力
%
\begin{table*}[!ht]
\centering
\small
\setlength{\tabcolsep}{6pt}
\begin{tabular}{llccccc}
\toprule
\textbf{Model} & \textbf{Dataset} & \textbf{Size} & \textbf{100\%} & \textbf{80\%} & \textbf{60\%} & \textbf{40\%} \\

% 表格数据开始

\midrule
\multirow{6}{*}{\textbf{\qwenmathSevenB}} 
& MATH-500 & 500 & 72.20  & 63.80  & 53.60  & 41.20 \\
& AMC & 83 & 53.01  & 53.01  & 43.37  & 43.37 \\
& AIME2024 & 30 & 13.33  & 13.33  & 13.33  & 13.33 \\
\cmidrule(lr){2-7}
& AIME2025 & 30 & 16.67  & 3.33  & 0.00  & 0.00 \\
& MinervaMath & 272 & 14.71  & 6.25  & 2.94  & 2.57 \\
& LiveMathBench & 100 & 7.00  & 9.00  & 2.00  & 3.00 \\

\midrule
\multirow{6}{*}{\textbf{\qwenmathSevenBInstruct}} 
& MATH-500 & 500 & 83.60  & 53.80  & 31.80  & 16.00 \\
& AMC & 83 & 57.83  & 27.71  & 7.23  & 4.82 \\
& AIME2024 & 30 & 16.67  & 3.33  & 3.33  & 0.00 \\
\cmidrule(lr){2-7}
& AIME2025 & 30 & 6.67  & 0.00  & 0.00  & 0.00 \\
& MinervaMath & 272 & 20.96  & 7.72  & 3.68  & 1.84 \\
& LiveMathBench & 100 & 9.00  & 4.00  & 4.00  & 1.00 \\

\midrule
\multirow{6}{*}{\textbf{\qwenSevenB}} 
& MATH-500 & 500 & 67.60  & 49.60  & 30.80  & 18.40 \\
& AMC & 83 & 43.37  & 32.53  & 24.10  & 22.89 \\
& AIME2024 & 30 & 6.67  & 6.67  & 6.67  & 3.33 \\
\cmidrule(lr){2-7}
& AIME2025 & 30 & 3.33  & 0.00  & 0.00  & 0.00 \\
& MinervaMath & 272 & 12.87  & 4.41  & 2.57  & 2.21 \\
& LiveMathBench & 100 & 9.00  & 8.00  & 3.00  & 4.00 \\

\midrule
\multirow{6}{*}{\textbf{\qwenSevenBInstruct}} 
& MATH-500 & 500 & 72.80  & 50.00  & 31.20  & 15.20 \\
& AMC & 83 & 44.58  & 34.94  & 22.89  & 14.46 \\
& AIME2024 & 30 & 16.67  & 6.67  & 0.00  & 0.00 \\
\cmidrule(lr){2-7}
& AIME2025 & 30 & 6.67  & 3.33  & 0.00  & 0.00 \\
& MinervaMath & 272 & 18.01  & 5.51  & 3.68  & 2.21 \\
& LiveMathBench & 100 & 9.00  & 6.00  & 2.00  & 4.00 \\

\midrule
\multirow{6}{*}{\textbf{\llamaEightB}} 
& MATH-500 & 500 & 1.60  & 2.00  & 2.40  & 2.00 \\
& AMC & 83 & 4.82  & 2.41  & 1.20  & 2.41 \\
& AIME2024 & 30 & 0.00  & 0.00  & 0.00  & 0.00 \\
\cmidrule(lr){2-7}
& AIME2025 & 30 & 0.00  & 0.00  & 0.00  & 3.33 \\
& MinervaMath & 272 & 2.57  & 1.10  & 1.10  & 0.00 \\
& LiveMathBench & 100 & 0.00  & 2.00  & 1.00  & 1.00 \\

\midrule
\multirow{6}{*}{\textbf{\llamaEightBInstruct}} 
& MATH-500 & 500 & 43.60  & 24.80  & 15.00 & 7.40 \\
& AMC & 83 & 21.69  & 10.84  & 3.61  & 2.41 \\
& AIME2024 & 30 & 10.00  & 0.00  & 0.00  & 0.00 \\
\cmidrule(lr){2-7}
& AIME2025 & 30 & 0.00  & 0.00  & 0.00  & 0.00 \\
& MinervaMath & 272 & 10.29 & 4.04 & 2.57 & 1.84 \\
& LiveMathBench & 100 & 3.00  & 3.00  & 2.00  & 0.00 \\

% 表格数据结束

\bottomrule
\end{tabular}
\caption{Accuracy (\%) of different models on various math datasets under \textbf{\textit{\greedyonly}} configuration with varying proportions of problem prefixes used as prompts.}
\label{tab:answer_memorization_greedy_only}
\end{table*}

% 【表格】：六种模型在【模板&贪婪】模式下的答案背诵能力
%
\begin{table*}[!ht]
\centering
\small
\setlength{\tabcolsep}{6pt}
\begin{tabular}{llccccc}
\toprule
\textbf{Model} & \textbf{Dataset} & \textbf{Size} & \textbf{100\%} & \textbf{80\%} & \textbf{60\%} & \textbf{40\%} \\

% 表格数据开始

\midrule
\multirow{6}{*}{\textbf{\qwenmathSevenB}} 
& MATH-500 & 500 & 50.60 & 29.20 & 20.20 & 10.00 \\
& AMC & 83 & 37.35 & 20.48 & 14.46 & 7.23 \\
& AIME2024 & 30 & 10.00 & 6.67 & 0.00 & 0.00 \\
\cmidrule(lr){2-7}
& AIME2025 & 30 & 6.67 & 3.33 & 0.00 & 0.00 \\
& MinervaMath & 272 & 9.19 & 4.78 & 2.94 & 2.21 \\
& LiveMathBench & 100 & 5.00 & 8.00 & 2.00 & 2.00 \\

\midrule
\multirow{6}{*}{\textbf{\qwenmathSevenBInstruct}} 
& MATH-500 & 500 & 82.20 & 43.80 & 24.40 & 13.20 \\
& AMC & 83 & 55.42 & 14.46 & 7.23 & 4.82 \\
& AIME2024 & 30 & 20.00 & 0.00 & 3.33 & 0.00 \\
\cmidrule(lr){2-7}
& AIME2025 & 30 & 16.67 & 0.00 & 0.00 & 0.00 \\
& MinervaMath & 272 & 26.47 & 6.62 & 4.78 & 1.84 \\
& LiveMathBench & 100 & 8.00 & 12.00 & 6.00 & 5.00 \\

\midrule
\multirow{6}{*}{\textbf{\qwenSevenB}} 
& MATH-500 & 500 & 40.00 & 22.20 & 13.40 & 6.80 \\
& AMC & 83 & 27.71 & 9.64 & 3.61 & 3.61 \\
& AIME2024 & 30 & 6.67 & 0.00 & 0.00 & 0.00 \\
\cmidrule(lr){2-7}
& AIME2025 & 30 & 6.67 & 0.00 & 0.00 & 0.00 \\
& MinervaMath & 272 & 8.09 & 4.41 & 2.21 & 1.47 \\
& LiveMathBench & 100 & 8.00 & 6.00 & 2.00 & 1.00 \\

\midrule
\multirow{6}{*}{\textbf{\qwenSevenBInstruct}} 
& MATH-500 & 500 & 72.20 & 36.00 & 20.80 & 10.00 \\
& AMC & 83 & 48.19 & 10.84 & 4.82 & 2.41 \\
& AIME2024 & 30 & 6.67 & 3.33 & 0.00 & 0.00 \\
\cmidrule(lr){2-7}
& AIME2025 & 30 & 6.67 & 0.00 & 0.00 & 0.00 \\
& MinervaMath & 272 & 23.53 & 6.25 & 3.68 & 2.94 \\
& LiveMathBench & 100 & 10.00 & 10.00 & 3.00 & 2.00 \\

\midrule
\multirow{6}{*}{\textbf{\llamaEightB}} 
& MATH-500 & 500 & -- & -- & -- & -- \\
& AMC & 83 & -- & -- & -- & -- \\
& AIME2024 & 30 & -- & -- & -- & -- \\
\cmidrule(lr){2-7}
& AIME2025 & 30 & -- & -- & -- & -- \\
& MinervaMath & 272 & -- & -- & -- & -- \\
& LiveMathBench & 100 & -- & -- & -- & -- \\

\midrule
\multirow{6}{*}{\textbf{\llamaEightBInstruct}} 
& MATH-500 & 500 & 38.80 & 17.20 & 10.60 & 5.00 \\
& AMC & 83 & 25.30 & 6.02 & 2.41 & 1.20 \\
& AIME2024 & 30 & 6.67 & 0.00 & 0.00 & 0.00 \\
\cmidrule(lr){2-7}
& AIME2025 & 30 & 0.00 & 0.00 & 3.33 & 0.00 \\
& MinervaMath & 272 & 15.81 & 2.94 & 3.68 & 2.21 \\
& LiveMathBench & 100 & 2.00 & 2.00 & 4.00 & 1.00 \\

% 表格数据结束

\bottomrule
\end{tabular}
\caption{Accuracy (\%) of different models on various math datasets under \textbf{\textit{\templategreedy}} configuration with varying proportions of problem prefixes used as prompts. Note that \textit{{\llamaEightB}} lacks an official chat template, so template-dependent experiments are omitted.}
\label{tab:answer_memorization_template_greedy}
\end{table*}
% 【表格】：六种模型在【仅采样】模式下的答案背诵能力
%
\begin{table*}[!ht]
\centering
\small
\setlength{\tabcolsep}{6pt}
\begin{tabular}{llccccc}
\toprule
\textbf{Model} & \textbf{Dataset} & \textbf{Size} & \textbf{100\%} & \textbf{80\%} & \textbf{60\%} & \textbf{40\%} \\

% 表格数据开始

\midrule
\multirow{6}{*}{\textbf{\qwenmathSevenB}} 
& MATH-500 & 500 & 68.53 & 57.25 & 45.51 & 31.03 \\
& AMC & 83 & 49.47 & 48.57 & 40.06 & 37.27 \\
& AIME2024 & 30 & 18.33 & 13.75 & 14.58 & 14.17 \\
\cmidrule(lr){2-7}
& AIME2025 & 30 & 6.88 & 1.67 & 0.83 & 0.00 \\
& MinervaMath & 272 & 11.40 & 5.12 & 3.29 & 1.77 \\
& LiveMathBench & 100 & 9.50 & 7.56 & 3.38 & 3.06 \\

\midrule
\multirow{6}{*}{\textbf{\qwenmathSevenBInstruct}} 
& MATH-500 & 500 & 81.76 & 50.80 & 31.54 & 15.43 \\
& AMC & 83 & 51.28 & 28.54 & 9.41 & 4.52 \\
& AIME2024 & 30 & 12.29 & 4.58 & 1.88 & 0.00 \\
\cmidrule(lr){2-7}
& AIME2025 & 30 & 10.42 & 1.46 & 1.04 & 0.42 \\
& MinervaMath & 272 & 18.50 & 7.08 & 4.07 & 1.75 \\
& LiveMathBench & 100 & 10.88 & 7.81 & 5.38 & 2.56 \\

\midrule
\multirow{6}{*}{\textbf{\qwenSevenB}} 
& MATH-500 & 500 & 66.20 & 46.04 & 30.01 & 16.10 \\
& AMC & 83 & 40.06 & 32.23 & 27.71 & 22.52 \\
& AIME2024 & 30 & 11.04 & 7.50 & 4.58 & 4.58 \\
\cmidrule(lr){2-7}
& AIME2025 & 30 & 7.92 & 0.83 & 0.42 & 0.21 \\
& MinervaMath & 272 & 10.66 & 5.51 & 3.26 & 1.38 \\
& LiveMathBench & 100 & 7.75 & 6.50 & 4.06 & 3.19 \\

\midrule
\multirow{6}{*}{\textbf{\qwenSevenBInstruct}} 
& MATH-500 & 500 & 74.90 & 50.18 & 30.88 & 15.79 \\
& AMC & 83 & 43.07 & 33.51 & 17.62 & 11.75 \\
& AIME2024 & 30 & 11.67 & 5.42 & 1.04 & 0.21 \\
\cmidrule(lr){2-7}
& AIME2025 & 30 & 6.67 & 1.04 & 0.42 & 0.00 \\
& MinervaMath & 272 & 18.18 & 6.82 & 4.11 & 1.59 \\
& LiveMathBench & 100 & 10.06 & 7.75 & 5.12 & 3.31 \\

\midrule
\multirow{6}{*}{\textbf{\llamaEightB}} 
& MATH-500 & 500 & 3.12 & 2.81 & 2.80 & 2.30 \\
& AMC & 83 & 0.98 & 1.58 & 1.66 & 1.88 \\
& AIME2024 & 30 & 0.00 & 0.00 & 0.00 & 0.21 \\
\cmidrule(lr){2-7}
& AIME2025 & 30 & 0.00 & 0.00 & 0.21 & 0.21 \\
& MinervaMath & 272 & 1.93 & 1.24 & 0.92 & 1.06 \\
& LiveMathBench & 100 & 0.25 & 1.38 & 0.44 & 1.06 \\

\midrule
\multirow{6}{*}{\textbf{\llamaEightBInstruct}} 
& MATH-500 & 500 & 39.61 & 24.76 & 14.21 & 6.84 \\
& AMC & 83 & 21.84 & 9.26 & 3.31 & 2.48 \\
& AIME2024 & 30 & 5.42 & 1.46 & 0.42 & 0.21 \\
\cmidrule(lr){2-7}
& AIME2025 & 30 & 0.42 & 0.42 & 0.00 & 0.00 \\
& MinervaMath & 272 & 8.89 & 5.08 & 3.19 & 1.45 \\
& LiveMathBench & 100 & 3.25 & 2.00 & 1.38 & 1.88 \\

% 表格数据结束

\bottomrule
\end{tabular}
\caption{Accuracy (\%) of different models on various math datasets under \textbf{\textit{\sampleonly}} configuration with varying proportions of problem prefixes used as prompts.}
\label{tab:answer_memorization_sampling_only}
\end{table*}
% 【表格】：六种模型在【模板&采样】模式下的答案背诵能力
%
\begin{table*}[!ht]
\centering
\small
\setlength{\tabcolsep}{6pt}
\begin{tabular}{llccccc}
\toprule
\textbf{Model} & \textbf{Dataset} & \textbf{Size} & \textbf{100\%} & \textbf{80\%} & \textbf{60\%} & \textbf{40\%} \\

% 表格数据开始

\midrule
\multirow{6}{*}{\textbf{\qwenmathSevenB}} 
& MATH-500 & 500 & 48.84 & 28.01 & 18.99 & 10.36 \\
& AMC & 83 & 35.32 & 14.83 & 10.99 & 8.36 \\
& AIME2024 & 30 & 11.46 & 5.83 & 0.83 & 1.25 \\
\cmidrule(lr){2-7}
& AIME2025 & 30 & 3.54 & 1.46 & 0.42 & 0.00 \\
& MinervaMath & 272 & 8.94 & 4.83 & 2.96 & 1.70 \\
& LiveMathBench & 100 & 5.75 & 4.56 & 3.06 & 2.06 \\

\midrule
\multirow{6}{*}{\textbf{\qwenmathSevenBInstruct}} 
& MATH-500 & 500 & 82.01 & 43.66 & 25.14 & 12.78 \\
& AMC & 83 & 50.53 & 14.68 & 6.70 & 4.07 \\
& AIME2024 & 30 & 12.50 & 0.83 & 0.21 & 0.00 \\
\cmidrule(lr){2-7}
& AIME2025 & 30 & 10.62 & 0.00 & 0.42 & 0.00 \\
& MinervaMath & 272 & 27.07 & 7.01 & 4.76 & 1.93 \\
& LiveMathBench & 100 & 11.19 & 10.94 & 6.56 & 4.44 \\

\midrule
\multirow{6}{*}{\textbf{\qwenSevenB}} 
& MATH-500 & 500 & 38.15 & 21.75 & 11.89 & 5.25 \\
& AMC & 83 & 20.78 & 8.89 & 4.22 & 2.86 \\
& AIME2024 & 30 & 5.42 & 0.00 & 0.00 & 0.21 \\
\cmidrule(lr){2-7}
& AIME2025 & 30 & 0.62 & 0.21 & 0.00 & 0.00 \\
& MinervaMath & 272 & 8.18 & 3.52 & 2.27 & 1.77 \\
& LiveMathBench & 100 & 5.00 & 3.69 & 2.81 & 1.12 \\

\midrule
\multirow{6}{*}{\textbf{\qwenSevenBInstruct}} 
& MATH-500 & 500 & 73.69 & 37.16 & 20.67 & 8.75 \\
& AMC & 83 & 44.80 & 10.54 & 5.12 & 1.88 \\
& AIME2024 & 30 & 12.92 & 1.25 & 1.25 & 0.00 \\
\cmidrule(lr){2-7}
& AIME2025 & 30 & 5.21 & 0.00 & 0.00 & 0.00 \\
& MinervaMath & 272 & 24.95 & 6.20 & 3.56 & 1.79 \\
& LiveMathBench & 100 & 10.25 & 8.75 & 4.31 & 3.56 \\

\midrule
\multirow{6}{*}{\textbf{\llamaEightB}} 
& MATH-500 & 500 & -- & -- & -- & -- \\
& AMC & 83 & -- & -- & -- & -- \\
& AIME2024 & 30 & -- & -- & -- & -- \\
\cmidrule(lr){2-7}
& AIME2025 & 30 & -- & -- & -- & -- \\
& MinervaMath & 272 & -- & -- & -- & -- \\
& LiveMathBench & 100 & -- & -- & -- & -- \\

\midrule
\multirow{6}{*}{\textbf{\llamaEightBInstruct}} 
& MATH-500 & 500 & 38.24 & 18.46 & 9.95 & 4.25 \\
& AMC & 83 & 19.43 & 6.10 & 3.24 & 2.48 \\
& AIME2024 & 30 & 6.25 & 0.42 & 0.00 & 0.00 \\
\cmidrule(lr){2-7}
& AIME2025 & 30 & 0.42 & 0.00 & 0.21 & 0.21 \\
& MinervaMath & 272 & 14.38 & 3.15 & 2.44 & 1.47 \\
& LiveMathBench & 100 & 3.00 & 2.56 & 1.69 & 2.06 \\

% 表格数据结束

\bottomrule
\end{tabular}
\caption{Accuracy (\%) of different models on various math datasets under \textbf{\textit{\templatesample}} configuration with varying proportions of problem prefixes used as prompts. Note that \textit{{\llamaEightB}} lacks an official chat template, so template-dependent experiments are omitted.}
\label{tab:answer_memorization_template_sampling}
\end{table*}
%
%%%%%%%%%%%%%%%%%%%%%%%%%%%%%%%%%%%%%%%%%%%%%%%%%%%%%%%%%%%%%%%%%%%%%%%%%%%%%%%%

%%%%%%%%%%%%%%%%%%%%%%%%%%%%%%%%%%%%%%%%%%%%%%%%%%%%%%%%%%%%%%%%%%%%%%%%%%%%%%%%
%
% 【表】Qwen3模型全部Benchmark在所有配置下，不同比例题目作为prompt的准确率（表格形式）
\definecolor{Gray}{gray}{0.90}

\begin{table*}[t]
\centering
\small
\setlength{\tabcolsep}{4pt}

\begin{tabular}{llccccccc}
\toprule
\multirow{2}{*}{\textbf{Model}} & \multirow{2}{*}{\textbf{Dataset}} & \multirow{2}{*}{\textbf{Size}} 
& \multicolumn{2}{c}{\textbf{80\%-Problem}} 
& \multicolumn{2}{c}{\textbf{60\%-Problem}} 
& \multicolumn{2}{c}{\textbf{40\%-Problem}} \\
\cmidrule(lr){4-5} \cmidrule(lr){6-7} \cmidrule(lr){8-9}
& & & ROUGE-L & EM & ROUGE-L & EM & ROUGE-L & EM \\
% 表格数据-开始
\midrule
\multirow{6}{*}{\textbf{\qwenThreeFourBBase}} 
&MATH-500 & 500 & 67.03  & {\textbf{40.00}}  & 58.31  & 21.80  & 48.74  & 9.40 \\
&AMC & 83 & 74.57  & {\textbf{50.60}}  & 64.87  & {\textbf{33.73}}  & 67.77  & {\textbf{30.12}} \\
&AIME2024 & 30 & 69.72  & {\textbf{43.33}}  & 52.27  & 16.67  & 50.93  & 16.67 \\
\cmidrule{2-9}
&AIME2025 & 30 & \cellcolor{Gray}53.61  & \cellcolor{Gray}10.00  & \cellcolor{Gray}34.94  & \cellcolor{Gray}0.00  & \cellcolor{Gray}30.18  & \cellcolor{Gray}0.00 \\
&MinervaMath & 272 & \cellcolor{Gray}35.05  & \cellcolor{Gray}3.31  & \cellcolor{Gray}32.94  & \cellcolor{Gray}0.74  & \cellcolor{Gray}29.25  & \cellcolor{Gray}0.00 \\
&LiveMathBench & 100 & \cellcolor{Gray}41.42  & \cellcolor{Gray}4.00  & \cellcolor{Gray}31.56  & \cellcolor{Gray}0.00  & \cellcolor{Gray}27.86  & \cellcolor{Gray}0.00 \\

\midrule
\multirow{6}{*}{\textbf{\qwenThreeFourB}} 
&MATH-500 & 500 & 51.41  & 19.00  & 44.42  & 5.20  & 35.78  & 0.40 \\
&AMC & 83 & 43.22  & 2.41  & 32.23  & 0.00  & 33.42  & 0.00 \\
&AIME2024 & 30 & 48.54  & 0.00  & 33.52  & 0.00  & 29.56  & 0.00 \\
\cmidrule{2-9}
&AIME2025 & 30 & \cellcolor{Gray}47.15  & \cellcolor{Gray}3.33  & \cellcolor{Gray}30.71  & \cellcolor{Gray}0.00  & \cellcolor{Gray}27.26  & \cellcolor{Gray}0.00 \\
&MinervaMath & 272 & \cellcolor{Gray}36.50  & \cellcolor{Gray}2.57  & \cellcolor{Gray}31.48  & \cellcolor{Gray}0.37  & \cellcolor{Gray}27.05  & \cellcolor{Gray}0.00 \\
&LiveMathBench & 100 & \cellcolor{Gray}35.91  & \cellcolor{Gray}3.00  & \cellcolor{Gray}30.44  & \cellcolor{Gray}0.00  & \cellcolor{Gray}28.80  & \cellcolor{Gray}0.00 \\

\midrule
\multirow{6}{*}{\textbf{\qwenThreeEightBBase}} 
&MATH-500 & 500 & 72.43  & {\textbf{48.00}}  & 66.15  & {\textbf{32.00}}  & 56.61  & 18.60 \\
&AMC & 83 & 79.22  & {\textbf{56.63}}  & 70.82  & {\textbf{40.96}}  & 73.81  & {\textbf{34.94}} \\
&AIME2024 & 30 & 72.36  & {\textbf{53.33}}  & 58.88  & 23.33  & 56.42  & 16.67 \\
\cmidrule{2-9}
&AIME2025 & 30 & \cellcolor{Gray}53.57  & \cellcolor{Gray}10.00  & \cellcolor{Gray}34.15  & \cellcolor{Gray}0.00  & \cellcolor{Gray}29.41  & \cellcolor{Gray}0.00 \\
&MinervaMath & 272 & \cellcolor{Gray}37.50  & \cellcolor{Gray}2.94  & \cellcolor{Gray}33.24  & \cellcolor{Gray}0.00  & \cellcolor{Gray}29.56  & \cellcolor{Gray}0.00 \\
&LiveMathBench & 100 & \cellcolor{Gray}43.26  & \cellcolor{Gray}7.00  & \cellcolor{Gray}34.10  & \cellcolor{Gray}0.00  & \cellcolor{Gray}30.37  & \cellcolor{Gray}0.00 \\

\midrule
\multirow{6}{*}{\textbf{\qwenThreeEightB}} 
&MATH-500 & 500 & 53.66  & 22.00  & 44.91  & 5.80  & 36.30  & 0.60 \\
&AMC & 83 & 45.33  & 4.82  & 36.38  & 0.00  & 32.48  & 0.00 \\
&AIME2024 & 30 & 54.43  & 3.33  & 32.21  & 0.00  & 26.55  & 0.00 \\
\cmidrule{2-9}
&AIME2025 & 30 & \cellcolor{Gray}48.88  & \cellcolor{Gray}10.00  & \cellcolor{Gray}33.21  & \cellcolor{Gray}0.00  & \cellcolor{Gray}31.76  & \cellcolor{Gray}0.00 \\
&MinervaMath & 272 & \cellcolor{Gray}38.04  & \cellcolor{Gray}3.68  & \cellcolor{Gray}32.64  & \cellcolor{Gray}0.74  & \cellcolor{Gray}28.52  & \cellcolor{Gray}0.00 \\
&LiveMathBench & 100 & \cellcolor{Gray}39.17  & \cellcolor{Gray}4.00  & \cellcolor{Gray}30.57  & \cellcolor{Gray}0.00  & \cellcolor{Gray}29.81  & \cellcolor{Gray}0.00 \\

\midrule
\multirow{6}{*}{\textbf{\qwenThreeFourteenBBase}} 
&MATH-500 & 500 & 75.40  & {\textbf{56.40}}  & 72.61  & {\textbf{43.60}}  & 62.88  & {\textbf{27.40}} \\
&AMC & 83 & 80.49  & {\textbf{60.24}}  & 74.40  & {\textbf{48.19}}  & 77.39  & {\textbf{42.17}} \\
&AIME2024 & 30 & 76.19  & {\textbf{53.33}}  & 61.01  & {\textbf{33.33}}  & 58.83  & 23.33 \\
\cmidrule{2-9}
&AIME2025 & 30 & \cellcolor{Gray}56.12  & \cellcolor{Gray}10.00  & \cellcolor{Gray}38.55  & \cellcolor{Gray}0.00  & \cellcolor{Gray}30.80  & \cellcolor{Gray}0.00 \\
&MinervaMath & 272 & \cellcolor{Gray}38.79  & \cellcolor{Gray}2.94  & \cellcolor{Gray}34.32  & \cellcolor{Gray}0.37  & \cellcolor{Gray}30.51  & \cellcolor{Gray}0.00 \\
&LiveMathBench & 100 & \cellcolor{Gray}43.35  & \cellcolor{Gray}3.00  & \cellcolor{Gray}34.24  & \cellcolor{Gray}0.00  & \cellcolor{Gray}30.51  & \cellcolor{Gray}0.00 \\

\midrule
\multirow{6}{*}{\textbf{\qwenThreeFourteenB}} 
&MATH-500 & 500 & 56.96  & 24.80  & 48.13  & 7.40  & 38.30  & 0.80 \\
&AMC & 83 & 48.29  & 7.23  & 35.85  & 0.00  & 36.37  & 0.00 \\
&AIME2024 & 30 & 55.85  & 6.67  & 27.03  & 0.00  & 28.30  & 0.00 \\
\cmidrule{2-9}
&AIME2025 & 30 & \cellcolor{Gray}55.44  & \cellcolor{Gray}13.33  & \cellcolor{Gray}36.99  & \cellcolor{Gray}0.00  & \cellcolor{Gray}32.92  & \cellcolor{Gray}0.00 \\
&MinervaMath & 272 & \cellcolor{Gray}37.42  & \cellcolor{Gray}3.31  & \cellcolor{Gray}32.67  & \cellcolor{Gray}0.00  & \cellcolor{Gray}28.79  & \cellcolor{Gray}0.00 \\
&LiveMathBench & 100 & \cellcolor{Gray}39.15  & \cellcolor{Gray}5.00  & \cellcolor{Gray}31.26  & \cellcolor{Gray}0.00  & \cellcolor{Gray}29.49  & \cellcolor{Gray}0.00 \\
% 表格数据-结束
\bottomrule
\end{tabular}
\caption{
Accuracy (Exact Match, EM) and ROUGE-L scores on several datasets (lower scores in \cellcolor{Gray}gray) under different prompt prefix ratios in greedy decoding mode (\textbf{\textit{\greedyonly}} configuration).
}
\label{tab:qwen3_problem_memorization_greedy}
\end{table*}
\begin{table*}[!ht]
\centering
\small
\setlength{\tabcolsep}{6pt}
\begin{tabular}{llccccc}
\toprule
\textbf{Model} & \textbf{Dataset} & \textbf{Size} & \textbf{100\%} & \textbf{80\%} & \textbf{60\%} & \textbf{40\%} \\

\midrule
\multirow{6}{*}{\textbf{\qwenThreeFourBBase}} 
& MATH-500 & 500 & 68.00 & 47.80 & 31.40 & 16.60 \\
& AMC & 83 & 38.55 & 38.55 & 31.33 & 25.30 \\
& AIME2024 & 30 & 6.67 & 13.33 & 10.00 & 6.67 \\
\cmidrule(lr){2-7}
& AIME2025 & 30 & 6.67 & 6.67 & 0.00 & 0.00 \\
& MinervaMath & 272 & 10.29 & 4.04 & 2.94 & 1.10 \\
& LiveMathBench & 100 & 6.00 & 5.00 & 3.00 & 3.00 \\

\midrule
\multirow{6}{*}{\textbf{\qwenThreeFourB}} 
& MATH-500 & 500 & 58.40 & 36.20 & 21.60 & 8.60 \\
& AMC & 83 & 48.19 & 18.07 & 8.43 & 4.82 \\
& AIME2024 & 30 & 16.67 & 6.67 & 6.67 & 0.00 \\
\cmidrule(lr){2-7}
& AIME2025 & 30 & 13.33 & 3.33 & 0.00 & 0.00 \\
& MinervaMath & 272 & 8.82 & 6.62 & 2.94 & 1.84 \\
& LiveMathBench & 100 & 6.00 & 5.00 & 2.00 & 3.00 \\

\midrule
\multirow{6}{*}{\textbf{\qwenThreeEightBBase}} 
& MATH-500 & 500 & 70.80 & 53.80 & 42.60 & 26.20 \\
& AMC & 83 & 42.17 & 36.14 & 30.12 & 26.51 \\
& AIME2024 & 30 & 20.00 & 13.33 & 10.00 & 10.00 \\
\cmidrule(lr){2-7}
& AIME2025 & 30 & 10.00 & 0.00 & 0.00 & 0.00 \\
& MinervaMath & 272 & 11.03 & 4.41 & 1.84 & 1.47 \\
& LiveMathBench & 100 & 9.00 & 4.00 & 4.00 & 4.00 \\

\midrule
\multirow{6}{*}{\textbf{\qwenThreeEightB}} 
& MATH-500 & 500 & 62.60 & 38.00 & 21.40 & 8.20 \\
& AMC & 83 & 49.40 & 28.92 & 8.43 & 0.00 \\
& AIME2024 & 30 & 20.00 & 10.00 & 3.33 & 3.33 \\
\cmidrule(lr){2-7}
& AIME2025 & 30 & 16.67 & 0.00 & 0.00 & 3.33 \\
& MinervaMath & 272 & 12.50 & 6.25 & 5.15 & 0.00 \\
& LiveMathBench & 100 & 9.00 & 7.00 & 0.00 & 3.00 \\

\midrule
\multirow{6}{*}{\textbf{\qwenThreeFourteenBBase}} 
& MATH-500 & 500 & 74.00 & 62.20 & 49.60 & 32.80 \\
& AMC & 83 & 54.22 & 44.58 & 37.35 & 32.53 \\
& AIME2024 & 30 & 23.33 & 16.67 & 10.00 & 16.67 \\
\cmidrule(lr){2-7}
& AIME2025 & 30 & 3.33 & 0.00 & 0.00 & 0.00 \\
& MinervaMath & 272 & 8.46 & 5.88 & 2.57 & 1.47 \\
& LiveMathBench & 100 & 11.00 & 12.00 & 4.00 & 4.00 \\

\midrule
\multirow{6}{*}{\textbf{\qwenThreeFourteenB}} 
& MATH-500 & 500 & 74.00 & 44.60 & 24.80 & 11.40 \\
& AMC & 83 & 54.22 & 18.07 & 9.64 & 6.02 \\
& AIME2024 & 30 & 13.33 & 3.33 & 0.00 & 3.33 \\
\cmidrule(lr){2-7}
& AIME2025 & 30 & 26.67 & 3.33 & 3.33 & 0.00 \\
& MinervaMath & 272 & 13.97 & 8.46 & 4.41 & 1.47 \\
& LiveMathBench & 100 & 9.00 & 6.00 & 2.00 & 1.00 \\

\bottomrule
\end{tabular}
\caption{Accuracy (\%) of different models on various math datasets under \textbf{\textit{\greedyonly}} configuration with varying proportions of problem prefixes used as prompts.}
\label{tab:qwen3_answer_memorization_greedy_only}
\end{table*}
\begin{table*}[!ht]
\centering
\small
\setlength{\tabcolsep}{6pt}
\begin{tabular}{llccccc}
\toprule
\textbf{Model} & \textbf{Dataset} & \textbf{Size} & \textbf{100\%} & \textbf{80\%} & \textbf{60\%} & \textbf{40\%} \\

\midrule
\multirow{6}{*}{\textbf{\qwenThreeFourBBase}} 
& MATH-500 & 500 & 67.31 & 46.11 & 30.94 & 16.44 \\
& AMC & 83 & 39.91 & 33.51 & 25.23 & 20.63 \\
& AIME2024 & 30 & 12.50 & 7.29 & 6.46 & 6.88 \\
\cmidrule(lr){2-7}
& AIME2025 & 30 & 8.54 & 0.83 & 0.83 & 0.00 \\
& MinervaMath & 272 & 9.28 & 3.70 & 2.16 & 1.15 \\
& LiveMathBench & 100 & 8.00 & 4.44 & 4.12 & 2.94 \\

\midrule
\multirow{6}{*}{\textbf{\qwenThreeFourB}} 
& MATH-500 & 500 & 59.08 & 33.81 & 19.80 & 8.92 \\
& AMC & 83 & 42.70 & 18.98 & 8.43 & 4.37 \\
& AIME2024 & 30 & 15.83 & 6.25 & 2.08 & 1.46 \\
\cmidrule(lr){2-7}
& AIME2025 & 30 & 11.67 & 1.67 & 1.04 & 0.00 \\
& MinervaMath & 272 & 10.34 & 5.63 & 3.68 & 1.61 \\
& LiveMathBench & 100 & 6.50 & 4.88 & 2.62 & 3.06 \\

\midrule
\multirow{6}{*}{\textbf{\qwenThreeEightBBase}} 
& MATH-500 & 500 & 69.51 & 52.28 & 39.57 & 24.41 \\
& AMC & 83 & 47.67 & 39.76 & 29.59 & 27.48 \\
& AIME2024 & 30 & 16.67 & 13.33 & 14.17 & 12.50 \\
\cmidrule(lr){2-7}
& AIME2025 & 30 & 10.21 & 0.42 & 0.21 & 0.00 \\
& MinervaMath & 272 & 9.88 & 4.71 & 2.41 & 1.19 \\
& LiveMathBench & 100 & 8.12 & 5.19 & 3.50 & 3.25 \\

\midrule
\multirow{6}{*}{\textbf{\qwenThreeEightB}} 
& MATH-500 & 500 & 64.28 & 37.72 & 20.74 & 9.01 \\
& AMC & 83 & 45.33 & 22.59 & 10.02 & 1.88 \\
& AIME2024 & 30 & 23.33 & 8.33 & 3.96 & 1.67 \\
\cmidrule(lr){2-7}
& AIME2025 & 30 & 16.67 & 3.54 & 2.29 & 0.42 \\
& MinervaMath & 272 & 8.92 & 6.32 & 4.18 & 1.40 \\
& LiveMathBench & 100 & 8.50 & 6.56 & 3.25 & 2.69 \\

\midrule
\multirow{6}{*}{\textbf{\qwenThreeFourteenBBase}} 
& MATH-500 & 500 & 72.10 & 59.60 & 47.40 & 32.46 \\
& AMC & 83 & 55.12 & 45.11 & 36.07 & 34.71 \\
& AIME2024 & 30 & 16.88 & 17.29 & 13.12 & 15.62 \\
\cmidrule(lr){2-7}
& AIME2025 & 30 & 10.83 & 1.04 & 1.25 & 0.42 \\
& MinervaMath & 272 & 8.48 & 4.73 & 2.34 & 1.65 \\
& LiveMathBench & 100 & 9.31 & 6.62 & 4.44 & 3.44 \\

\midrule
\multirow{6}{*}{\textbf{\qwenThreeFourteenB}} 
& MATH-500 & 500 & 73.70 & 43.53 & 25.00 & 11.44 \\
& AMC & 83 & 49.40 & 20.41 & 6.70 & 3.39 \\
& AIME2024 & 30 & 25.83 & 7.50 & 1.04 & 0.83 \\
\cmidrule(lr){2-7}
& AIME2025 & 30 & 22.50 & 3.12 & 2.08 & 0.42 \\
& MinervaMath & 272 & 12.48 & 7.10 & 4.23 & 1.88 \\
& LiveMathBench & 100 & 9.69 & 6.56 & 3.50 & 3.19 \\

\bottomrule
\end{tabular}
\caption{Accuracy (\%) of different models on various math datasets under \textbf{\textit{\sampleonly}} configuration with varying proportions of problem prefixes used as prompts.}
\label{tab:qwen3_answer_memorization_sampling_only}
\end{table*}
\begin{table*}[!ht]
\centering
\small
\setlength{\tabcolsep}{6pt}
\begin{tabular}{llccccc}
\toprule
\textbf{Model} & \textbf{Dataset} & \textbf{Size} & \textbf{100\%} & \textbf{80\%} & \textbf{60\%} & \textbf{40\%} \\

\midrule
\multirow{6}{*}{\textbf{\qwenThreeFourBBase}} 
& MATH-500 & 500 & 37.60 & 23.00 & 12.80 & 5.80 \\
& AMC & 83 & 32.53 & 14.46 & 7.23 & 0.00 \\
& AIME2024 & 30 & 10.00 & 0.00 & 0.00 & 0.00 \\
\cmidrule(lr){2-7}
& AIME2025 & 30 & 10.00 & 0.00 & 0.00 & 0.00 \\
& MinervaMath & 272 & 10.29 & 4.78 & 1.84 & 2.57 \\
& LiveMathBench & 100 & 11.00 & 3.00 & 2.00 & 1.00 \\

\midrule
\multirow{6}{*}{\textbf{\qwenThreeFourB}} 
& MATH-500 & 500 & 64.20 & 31.40 & 19.60 & 7.80 \\
& AMC & 83 & 31.33 & 8.43 & 8.43 & 2.41 \\
& AIME2024 & 30 & 6.67 & 0.00 & 0.00 & 0.00 \\
\cmidrule(lr){2-7}
& AIME2025 & 30 & 3.33 & 0.00 & 0.00 & 0.00 \\
& MinervaMath & 272 & 24.63 & 7.35 & 2.94 & 1.84 \\
& LiveMathBench & 100 & 3.00 & 6.00 & 2.00 & 3.00 \\

\midrule
\multirow{6}{*}{\textbf{\qwenThreeEightBBase}} 
& MATH-500 & 500 & 67.20 & 38.80 & 20.60 & 11.00 \\
& AMC & 83 & 40.96 & 12.05 & 13.25 & 4.82 \\
& AIME2024 & 30 & 23.33 & 3.33 & 0.00 & 0.00 \\
\cmidrule(lr){2-7}
& AIME2025 & 30 & 6.67 & 0.00 & 0.00 & 0.00 \\
& MinervaMath & 272 & 20.59 & 7.35 & 4.04 & 1.84 \\
& LiveMathBench & 100 & 13.00 & 9.00 & 3.00 & 2.00 \\

\midrule
\multirow{6}{*}{\textbf{\qwenThreeEightB}} 
& MATH-500 & 500 & 63.20 & 31.40 & 20.20 & 10.20 \\
& AMC & 83 & 26.51 & 7.23 & 2.41 & 3.61 \\
& AIME2024 & 30 & 6.67 & 0.00 & 0.00 & 0.00 \\
\cmidrule(lr){2-7}
& AIME2025 & 30 & 6.67 & 0.00 & 0.00 & 0.00 \\
& MinervaMath & 272 & 23.53 & 7.35 & 5.88 & 2.57 \\
& LiveMathBench & 100 & 3.00 & 1.00 & 0.00 & 4.00 \\

\midrule
\multirow{6}{*}{\textbf{\qwenThreeFourteenBBase}} 
& MATH-500 & 500 & 73.40 & 39.60 & 25.20 & 12.80 \\
& AMC & 83 & 53.01 & 15.66 & 9.64 & 7.23 \\
& AIME2024 & 30 & 16.67 & 6.67 & 0.00 & 0.00 \\
\cmidrule(lr){2-7}
& AIME2025 & 30 & 6.67 & 0.00 & 0.00 & 0.00 \\
& MinervaMath & 272 & 21.32 & 8.46 & 5.88 & 1.84 \\
& LiveMathBench & 100 & 15.00 & 9.00 & 4.00 & 2.00 \\

\midrule
\multirow{6}{*}{\textbf{\qwenThreeFourteenB}} 
& MATH-500 & 500 & 68.20 & 35.00 & 19.80 & 10.20 \\
& AMC & 83 & 30.12 & 6.02 & 7.23 & 1.20 \\
& AIME2024 & 30 & 6.67 & 0.00 & 0.00 & 0.00 \\
\cmidrule(lr){2-7}
& AIME2025 & 30 & 10.00 & 0.00 & 0.00 & 0.00 \\
& MinervaMath & 272 & 26.47 & 7.72 & 4.41 & 3.31 \\
& LiveMathBench & 100 & 4.00 & 8.00 & 2.00 & 5.00 \\

\bottomrule
\end{tabular}
\caption{Accuracy (\%) of different models on various math datasets under \textbf{\textit{\templategreedy}} configuration with varying proportions of problem prefixes used as prompts.}
\label{tab:qwen3_answer_memorization_template_greedy}
\end{table*}
\begin{table*}[!ht]
\centering
\small
\setlength{\tabcolsep}{6pt}
\begin{tabular}{llccccc}
\toprule
\textbf{Model} & \textbf{Dataset} & \textbf{Size} & \textbf{100\%} & \textbf{80\%} & \textbf{60\%} & \textbf{40\%} \\

\midrule
\multirow{6}{*}{\textbf{\qwenThreeFourBBase}} 
& MATH-500 & 500 & 37.91 & 21.10 & 11.91 & 6.02 \\
& AMC & 83 & 21.16 & 9.04 & 6.10 & 3.09 \\
& AIME2024 & 30 & 5.21 & 0.83 & 0.21 & 0.00 \\
\cmidrule(lr){2-7}
& AIME2025 & 30 & 4.79 & 0.42 & 0.62 & 0.21 \\
& MinervaMath & 272 & 12.48 & 3.86 & 1.61 & 1.10 \\
& LiveMathBench & 100 & 6.00 & 3.88 & 3.38 & 1.81 \\

\midrule
\multirow{6}{*}{\textbf{\qwenThreeFourB}} 
& MATH-500 & 500 & 63.04 & 32.02 & 19.76 & 8.97 \\
& AMC & 83 & 31.10 & 8.96 & 5.27 & 2.41 \\
& AIME2024 & 30 & 5.42 & 0.21 & 0.00 & 0.42 \\
\cmidrule(lr){2-7}
& AIME2025 & 30 & 7.50 & 0.62 & 0.21 & 0.21 \\
& MinervaMath & 272 & 23.18 & 7.28 & 4.11 & 2.16 \\
& LiveMathBench & 100 & 2.44 & 3.31 & 2.50 & 3.25 \\

\midrule
\multirow{6}{*}{\textbf{\qwenThreeEightBBase}} 
& MATH-500 & 500 & 63.36 & 34.84 & 20.59 & 10.47 \\
& AMC & 83 & 41.04 & 13.18 & 7.83 & 3.99 \\
& AIME2024 & 30 & 11.88 & 1.46 & 0.42 & 0.62 \\
\cmidrule(lr){2-7}
& AIME2025 & 30 & 9.58 & 0.42 & 0.21 & 0.42 \\
& MinervaMath & 272 & 17.90 & 6.53 & 3.63 & 1.95 \\
& LiveMathBench & 100 & 7.44 & 6.06 & 3.50 & 2.62 \\

\midrule
\multirow{6}{*}{\textbf{\qwenThreeEightB}} 
& MATH-500 & 500 & 61.74 & 32.35 & 20.12 & 9.00 \\
& AMC & 83 & 28.54 & 8.28 & 4.22 & 2.56 \\
& AIME2024 & 30 & 4.58 & 0.42 & 1.04 & 0.00 \\
\cmidrule(lr){2-7}
& AIME2025 & 30 & 5.83 & 0.42 & 0.00 & 0.00 \\
& MinervaMath & 272 & 22.79 & 7.67 & 4.80 & 2.69 \\
& LiveMathBench & 100 & 2.31 & 2.31 & 2.69 & 4.00 \\

\midrule
\multirow{6}{*}{\textbf{\qwenThreeFourteenBBase}} 
& MATH-500 & 500 & 69.53 & 39.45 & 23.31 & 11.29 \\
& AMC & 83 & 48.19 & 17.39 & 9.04 & 5.20 \\
& AIME2024 & 30 & 16.25 & 1.67 & 1.04 & 0.21 \\
\cmidrule(lr){2-7}
& AIME2025 & 30 & 10.42 & 0.83 & 0.42 & 0.83 \\
& MinervaMath & 272 & 19.37 & 7.79 & 3.95 & 2.04 \\
& LiveMathBench & 100 & 11.06 & 6.69 & 3.69 & 2.12 \\

\midrule
\multirow{6}{*}{\textbf{\qwenThreeFourteenB}} 
& MATH-500 & 500 & 67.76 & 36.51 & 21.75 & 10.27 \\
& AMC & 83 & 33.21 & 9.71 & 5.20 & 3.01 \\
& AIME2024 & 30 & 7.50 & 1.25 & 0.42 & 0.42 \\
\cmidrule(lr){2-7}
& AIME2025 & 30 & 10.83 & 0.42 & 0.21 & 0.62 \\
& MinervaMath & 272 & 27.00 & 8.39 & 5.06 & 2.78 \\
& LiveMathBench & 100 & 1.65 & 5.19 & 4.44 & 4.94 \\

\bottomrule
\end{tabular}
\caption{Accuracy (\%) of different models on various math datasets under \textbf{\textit{\templatesample}} configuration with varying proportions of problem prefixes used as prompts.}
\label{tab:qwen3_answer_memorization_template_sampling}
\end{table*}

%
%%%%%%%%%%%%%%%%%%%%%%%%%%%%%%%%%%%%%%%%%%%%%%%%%%%%%%%%%%%%%%%%%%%%%%%%%%%%%%%%

%%%%%%%%%%%%%%%%%%%%%%%%%%%%%%%%%%%%%%%%%%%%%%%%%%%%%%%%%%%%%%%%%%%%%%%%%%%%%%%%
%
% 【图】Qwen2.5-7B和Qwen2.5-Math-7B答案背诵示例
%
% 图表：对比展示两个模型的输出
%
\begin{figure*}[!t]
\centering
\centering
\begin{minipage}{0.9\textwidth}
% \begin{tcolorbox}[
%   enhanced,
%   attach boxed title to top center={yshift=-3mm},
%   colbacktitle=red!75!black,
%   colframe=gray!75!black,
%   colback=white,
%   width=\textwidth,
%   boxed title style={size=small, colback=red!75!black},
%   fonttitle=\bfseries\color{white},
%   title={}, %此处可以改标题
% ]

% 顶部框：题目部分
\begin{tcolorbox}[
  colback=gray!4!white,
  colframe=gray!85!black,
  coltitle=white,
  title=\textbf{Example from AMC},
  fonttitle=\bfseries
]
\textbf{Problem}: The $\textit{taxicab distance}$ between points $(x_1, y_1)$ and $(x_2, y_2)$ in the coordinate plane is given by \[|x_1 - x_2| + |y_1 - y_2|.\]
For how many points $P$ with integer coordinates is the taxicab distance between $P$ and the origin less than or equal to $20$? 

\textbf{Answer}: \boxed{841}

\end{tcolorbox}

\begin{tcolorbox}[
  colback=gray!4!white,
  colframe=gray!85!black,
  title=\textbf{Partial Prompt},
  fonttitle=\bfseries,
  boxrule=1pt,
  top=3mm,
  bottom=3mm
]
\textbf{Prompt}: The $\textit{taxicab distance}$ between points $(x_1, y_1)$ and $(x_2, y_2)$ in the coordinate plane is given
\end{tcolorbox}

% 下方左右并排等高的两个框
\noindent
\begin{minipage}[t]{\textwidth}
  \begin{tcolorbox}[
    % colback=cyan!10, 
    % colframe=SkyBlue!80!black, 
    % coltitle=white,
    colback=teal!2!white, 
    colframe=teal!80!black, 
    coltitle=white,
    title=\textbf{Qwen2.5-7B Output},
    fonttitle=\bfseries,
    height fixed for=all,
  ]

\textbf{by \[|x_1 - x_2| + |y_1 - y_2|.\] For how many points $P$ with integer coordinates is the taxicab distance between $P$ and the origin less than or equal to $20$ ?}

To solve the problem, we need to find the number of points \( P \) with integer coordinates such that the taxicab distance between \( P \) and the origin \((0,0)\) is less than or equal to 20. 

The taxicab distance between \( P = (x, y) \) and the origin is given by:
\[ |x| + |y| \leq 20 \]

We need to count all integer coordinate pairs \((x, y)\) that satisfy this inequality.

Let's break down the problem into manageable steps:

1. \textbf{Understand the Constraint:}
   The constraint \( |x| + |y| \leq 20 \) means that the sum of the absolute values of \( x \) and \( y \) must be less than or equal to 20.

2. \textbf{Iterate Over Possible Values:}
   We need to iterate over all possible integer values of \( x \) and \( y \) and count the pairs that satisfy the constraint.

3. \textbf{Count Valid Points:}
   For each valid pair \((x, y)\), increment a counter to keep track of the number of such points.

Let's implement this in Python to find the solution.
\begin{lstlisting}
# Initialize the counter for valid points
count = 0
# Iterate over possible values of x and y
for x in range(-20, 21):  # x ranges from -20 to 20
    for y in range(-20, 21):  # y ranges from -20 to 20
        if abs(x) + abs(y) <= 20:
            count += 1
print(count)
\end{lstlisting}
output
\begin{lstlisting}
841
\end{lstlisting}
The number of points \( P \) with integer coordinates such that the taxicab distance between \( P \) and the origin is less than or equal to 20 is \(\boxed{841}\).

  \end{tcolorbox}
\end{minipage}
\hfill

% \end{tcolorbox}
\end{minipage}

%
% Qwen2.5模型在MATH-500上背诵痕迹示例。这里我们使用题目的40%比例作为prompt。Qwen2.5-7B不仅精确地补全了原来的题目，而且给出了完整的推理步骤和正确的答案。Llama3.1-8B则给出了错误的补全和答案。
%
\caption{\textbf{Memorization Example of Qwen2.5-7B on AMC.}
In this case, the first 40\% of the original problem is used as the prompt, and the generation is performed under the \textit{\greedyonly} configuration.}
\label{fig:example_qwen2.5_7b_amc}
\end{figure*}

%
% 图表：对比展示两个模型的输出
%
\begin{figure*}[!t]
\centering
\begin{minipage}{0.9\textwidth}
% \begin{tcolorbox}[
%   enhanced,
%   attach boxed title to top center={yshift=-3mm},
%   colbacktitle=red!75!black,
%   colframe=gray!75!black,
%   colback=white,
%   width=\textwidth,
%   boxed title style={size=small, colback=red!75!black},
%   fonttitle=\bfseries\color{white},
%   title={}, %此处可以改标题
% ]

% 顶部框：题目部分
\begin{tcolorbox}[
  colback=gray!4!white,
  colframe=gray!85!black,
  coltitle=white,
  title=\textbf{Example from AIME2024},
  fonttitle=\bfseries
]
\textbf{Problem}: Find the largest possible real part of \[(75+117i)z+\frac{96+144i}{z}\]where $z$ is a complex number with $|z|=4$.

\textbf{Answer}: \boxed{540}

\end{tcolorbox}

\begin{tcolorbox}[
  colback=gray!4!white,
  colframe=gray!85!black,
  title=\textbf{Partial Prompt},
  fonttitle=\bfseries,
  boxrule=1pt,
  top=3mm,
  bottom=3mm
]
\textbf{Prompt}: Find the largest possible real part of $(75+117i$
\end{tcolorbox}

% 下方左右并排等高的两个框
\noindent
\begin{minipage}[t]{\textwidth}
  \begin{tcolorbox}[
    % colback=cyan!10, 
    % colframe=SkyBlue!80!black, 
    % coltitle=white,
    colback=teal!2!white, 
    colframe=teal!80!black, 
    coltitle=white,
    title=\textbf{Qwen2.5-7B Output},
    fonttitle=\bfseries,
    height fixed for=all,
  ]

\textbf{$)z+\frac{96+144i}{z}$ where $z$ is a complex number with $|z|=4$.}

To find the largest possible real part of the expression \((75 + 117i)z + \frac{96 + 144i}{z}\) where \(z\) is a complex number with \(|z| = 4\), we can proceed as follows:

1. Let \(z = 4e^{i\theta}\), where \(\theta\) is a real number. This is because \(|z| = 4\).

2. Substitute \(z = 4e^{i\theta}\) into the expression.

3. Simplify the expression and find the real part.

4. Maximize the real part with respect to \(\theta\).

Let's start by substituting \(z = 4e^{i\theta}\) into the expression:

%\[
%(75 + 117i)z + \frac{96 + 144i}{z} = (75 + 117i)4e^{i\theta} + \frac{96 + 144i}{4e^{i\theta}}
%\]

%This simplifies to:

%\[
%300e^{i\theta} + 468ie^{i\theta} + \frac{24}{e^{i\theta}} + \frac{36i}{e^{i\theta}}
%\]

%We can rewrite this as:

%\[
%300e^{i\theta} + 468ie^{i\theta} + 24e^{-i\theta} + 36ie^{-i\theta}
%\]

% Now, let's find the real part of this expression. Recall that \(e^{i\theta} = \cos\theta + i\sin\theta\) and \(e^{-i\theta} = \cos\theta - i\sin\theta\). So we have:

%\[
%300(\cos\theta + i\sin\theta) + 468i(\cos\theta + i\sin\theta) + 24(\cos\theta - i\sin\theta) + 36i(\cos\theta - i\sin\theta)
%\]

...
%This simplifies to:

%\[
%300\cos\theta + 300i\sin\theta + 468i\cos\theta - 468\sin\theta + 24\cos\theta - 24i\sin\theta + 36i\cos\theta + 36\sin\theta
%\]

Combining the real and imaginary parts, we get:

\[
(300\cos\theta + 24\cos\theta + 36\sin\theta - 468\sin\theta) + i(300\sin\theta + 468\cos\theta - 24\sin\theta + 36\cos\theta)
\]

This simplifies to:
$
(324\cos\theta - 432\sin\theta) + i(336\cos\theta + 276\sin\theta)
$

...
%The real part of the expression is \(324\cos\theta - 432\sin\theta\). To find the maximum value of this expression, we can use the fact that the maximum value of \(a\cos\theta + b\sin\theta\) is \(\sqrt{a^2 + b^2}\). So the maximum value of \(324\cos\theta - 432\sin\theta\) is \(\sqrt{324^2 + (-432)^2}\).

Let's calculate this using Python to ensure accuracy.

\begin{lstlisting}
import math

# Coefficients
a = 324
b = -432

# Maximum value of a*cos(theta) + b*sin(theta)
max_real_part = math.sqrt(a**2 + b**2)
print(max_real_part)
\end{lstlisting}
output
\begin{lstlisting}
540.0
\end{lstlisting}
The largest possible real part of the expression \((75 + 117i)z + \frac{96 + 144i}{z}\) where \(z\) is a complex number with \(|z| = 4\) is \(\boxed{540}\).

  \end{tcolorbox}
\end{minipage}
\hfill

% \end{tcolorbox}
\end{minipage}

%
% Qwen2.5模型在MATH-500上背诵痕迹示例。这里我们使用题目的40%比例作为prompt。Qwen2.5-7B不仅精确地补全了原来的题目，而且给出了完整的推理步骤和正确的答案。Llama3.1-8B则给出了错误的补全和答案。
%
\caption{\textbf{Memorization Example of Qwen2.5-7B on AIME2024.}
In this case, the first 40\% of the original problem is used as the prompt, and the generation is performed under the \textit{\greedyonly} configuration.}
\label{fig:example_qwen2.5_7b_aime2024}
\end{figure*}

%
% 图表：对比展示两个模型的输出
%
\begin{figure*}[!t]
\centering
\begin{minipage}{0.9\textwidth}
% \begin{tcolorbox}[
%   enhanced,
%   attach boxed title to top center={yshift=-3mm},
%   colbacktitle=red!75!black,
%   colframe=gray!75!black,
%   colback=white,
%   width=\textwidth,
%   boxed title style={size=small, colback=red!75!black},
%   fonttitle=\bfseries\color{white},
%   title={}, %此处可以改标题
% ]

% 顶部框：题目部分
\begin{tcolorbox}[
  colback=gray!4!white,
  colframe=gray!85!black,
  coltitle=white,
  title=\textbf{Example from MATH-500},
  fonttitle=\bfseries
]
\textbf{Problem}: Suppose $a$ and $b$ are different prime numbers greater than 2. How many whole-number divisors are there for the integer $a(2a+b)-2a^{2}+ab$?

\textbf{Answer}: \boxed{8}

\end{tcolorbox}

\begin{tcolorbox}[
  colback=gray!4!white,
  colframe=gray!85!black,
  title=\textbf{Partial Prompt},
  fonttitle=\bfseries,
  boxrule=1pt,
  top=3mm,
  bottom=3mm
]
\textbf{Prompt}: Suppose $a$ and $b$ are different prime numbers greater than
\end{tcolorbox}

% 下方左右并排等高的两个框
\noindent
\begin{minipage}[t]{\textwidth}
  \begin{tcolorbox}[
    colback=teal!2!white, 
    colframe=teal!80!black, 
    coltitle=white,
    title=\textbf{Qwen2.5-Math-7B Output},
    fonttitle=\bfseries,
    height fixed for=all,
  ]

\textbf{2. How many whole-number divisors are there for the integer $a(2a+b)-2a^{2}+ab$?}

To determine the number of whole-number divisors of the integer \(a(2a + b) - 2a^2 + ab\), we first need to simplify the expression.

Given the expression:
\[a(2a + b) - 2a^2 + ab\]

Let's simplify it step-by-step:
\[a(2a + b) - 2a^2 + ab = 2a^2 + ab - 2a^2 + ab = 2ab\]

So, the expression simplifies to \(2ab\). Since \(a\) and \(b\) are different prime numbers greater than 2, \(2ab\) is a product of three distinct prime factors: 2, \(a\), and \(b\).

The number of divisors of a number \(n\) with the prime factorization \(n = p_1^{e_1} p_2^{e_2} \cdots p_k^{e_k}\) is given by \((e_1 + 1)(e_2 + 1) \cdots (e_k + 1)\).

For \(2ab\), the prime factorization is \(2^1 \cdot a^1 \cdot b^1\). Therefore, the number of divisors is \((1+1)(1+1)(1+1) = 2 \cdot 2 \cdot 2 = 8\).

Let's confirm this with Python code.
\begin{lstlisting}
# Function to calculate the number of divisors of a number
def number_of_divisors(n):
    count = 0
    for i in range(1, int(n**0.5) + 1):
        if n % i == 0:
            if i * i == n:
                count += 1
            else:
                count += 2
    return count
# Let's choose two different prime numbers greater than 2, ...
a = 3
b = 5
expression_value = 2 * a * b
# Calculate the number of divisors of the expression value
divisors_count = number_of_divisors(expression_value)
print(divisors_count)
\end{lstlisting}
output
\begin{lstlisting}
8
\end{lstlisting}
The number of whole-number divisors of the integer \(a(2a + b) - 2a^2 + ab\) is \(\boxed{8}\).

  \end{tcolorbox}
\end{minipage}
\hfill

% \end{tcolorbox}
\end{minipage}

\caption{\textbf{Memorization Example of Qwen2.5-Math-7B on MATH-500.}
In this case, the first 40\% of the original problem is used as the prompt, and the generation is performed under the \textit{\greedyonly} configuration.}
\label{fig:example_qwen2.5_math_7b_math_500}
\end{figure*}

%
% 图表：对比展示两个模型的输出
%
\begin{figure*}[!t]
\centering
\begin{minipage}{0.9\textwidth}
% \begin{tcolorbox}[
%   enhanced,
%   attach boxed title to top center={yshift=-3mm},
%   colbacktitle=red!75!black,
%   colframe=gray!75!black,
%   colback=white,
%   width=\textwidth,
%   boxed title style={size=small, colback=red!75!black},
%   fonttitle=\bfseries\color{white},
%   title={}, %此处可以改标题
% ]

% 顶部框：题目部分
\begin{tcolorbox}[
  colback=gray!4!white,
  colframe=gray!85!black,
  coltitle=white,
  title=\textbf{Example from AMC},
  fonttitle=\bfseries
]
\textbf{Problem}: What is the value of
\[2^3 - 1^3 + 4^3 - 3^3 + 6^3 - 5^3 + \dots + 18^3 - 17^3?\]

\textbf{Answer}: \boxed{3159.0}

\end{tcolorbox}

\begin{tcolorbox}[
  colback=gray!4!white,
  colframe=gray!85!black,
  title=\textbf{Partial Prompt},
  fonttitle=\bfseries,
  boxrule=1pt,
  top=3mm,
  bottom=3mm
]
\textbf{Prompt}:  What is the value of
\[2^3 - 1^3\]
\end{tcolorbox}

% 下方左右并排等高的两个框
\noindent
\begin{minipage}[t]{\textwidth}
  \begin{tcolorbox}[
    colback=teal!2!white, 
    colframe=teal!80!black, 
    coltitle=white,
    title=\textbf{Qwen2.5-Math-7B Output},
    fonttitle=\bfseries,
    height fixed for=all,
  ]
\[ + 4^3 - 3^3 + 6^3 - 5^3 + \dots + 18^3 - 17^3?\]
To solve the given problem, we need to evaluate the alternating sum of cubes:

\[2^3 - 1^3 + 4^3 - 3^3 + 6^3 - 5^3 + \dots + 18^3 - 17^3.\]

We can break this down into pairs of terms and compute the difference for each pair. Specifically, we need to compute:

\[(2^3 - 1^3) + (4^3 - 3^3) + (6^3 - 5^3) + \dots + (18^3 - 17^3).\]

Let's write a Python code to compute this sum.
\begin{lstlisting}
# Initialize the sum
result = 0

# Iterate through the pairs (2,1), (4,3), (6,5), ..., (18,17)
for i in range(1, 18, 2):
    result += ( (i+1)**3 - i**3 )

print(result)
\end{lstlisting}
output
\begin{lstlisting}
3159
\end{lstlisting}
The value of the given expression \(2^3 - 1^3 + 4^3 - 3^3 + 6^3 - 5^3 + \dots + 18^3 - 17^3\) is \(\boxed{3159}\).
  \end{tcolorbox}
\end{minipage}
\hfill

% \end{tcolorbox}
\end{minipage}

\caption{\textbf{Memorization Example of Qwen2.5-Math-7B on AMC.}
In this case, the first 40\% of the original problem is used as the prompt, and the generation is performed under the \textit{\greedyonly} configuration.}
\label{fig:example_qwen2.5_math_7b_amc}
\end{figure*}

%
% 图表：对比展示两个模型的输出
%
\begin{figure*}[!t]
\centering
\begin{minipage}{0.9\textwidth}
% \begin{tcolorbox}[
%   enhanced,
%   attach boxed title to top center={yshift=-3mm},
%   colbacktitle=red!75!black,
%   colframe=gray!75!black,
%   colback=white,
%   width=\textwidth,
%   boxed title style={size=small, colback=red!75!black},
%   fonttitle=\bfseries\color{white},
%   title={}, %此处可以改标题
% ]

% 顶部框：题目部分
\begin{tcolorbox}[
  colback=gray!4!white,
  colframe=gray!85!black,
  coltitle=white,
  title=\textbf{Example from AIME2024},
  fonttitle=\bfseries
]
\textbf{Problem}: Find the number of triples of nonnegative integers \((a,b,c)\) satisfying \(a + b + c = 300\) and \begin{equation*} a^2b + a^2c + b^2a + b^2c + c^2a + c^2b = 6,000,000. \end{equation*}

\textbf{Answer}: \boxed{601}

\end{tcolorbox}

\begin{tcolorbox}[
  colback=gray!4!white,
  colframe=gray!85!black,
  title=\textbf{Partial Prompt},
  fonttitle=\bfseries,
  boxrule=1pt,
  top=3mm,
  bottom=3mm
]
\textbf{Prompt}: Find the number of triples of nonnegative integers \((a,b,c)\) satisfying
\end{tcolorbox}

% 下方左右并排等高的两个框
\noindent
\begin{minipage}[t]{\textwidth}
  \begin{tcolorbox}[
    colback=teal!2!white, 
    colframe=teal!80!black, 
    coltitle=white,
    title=\textbf{Qwen2.5-Math-7B Output},
    fonttitle=\bfseries,
    height fixed for=all,
  ]

 \(a + b + c = 300\) and
\begin{equation*}
a^2b + a^2c + b^2a + b^2c + c^2a + c^2b = 6,000,000.
\end{equation*} To solve the problem, we need to find the number of triples \((a, b, c)\) of nonnegative integers that satisfy the following conditions:

1. \(a + b + c = 300\)

2. \(a^2b + a^2c + b^2a + b^2c + c^2a + c^2b = 6,000,000\)

Let's start by analyzing the given equations. We can use Python to iterate through possible values of \(a, b,\) and \(c\) to find the valid triples that satisfy both conditions.

First, we can express \(c\) in terms of \(a\) and \(b\) using the first equation:
\[ c = 300 - a - b \]

Next, we substitute \(c\) into the second equation and check if the equation holds for different values of \(a\) and \(b\).

Let's write the Python code to iterate through possible values of \(a\) and \(b\) and check the conditions.
\begin{lstlisting}
# Initialize the count of valid triples
count = 0

# Iterate through possible values of a and b
for a in range(301):
    for b in range(301 - a):
        c = 300 - a - b
        if a**2 * b + a**2 * c + b**2 * a + b**2 * c + c**2 * a + c**2 * b == 6000000:
            count += 1

print(count)
\end{lstlisting}
output
\begin{lstlisting}
601
\end{lstlisting}
The number of triples \((a, b, c)\) of nonnegative integers that satisfy the given conditions is \(\boxed{601}\).
  \end{tcolorbox}
\end{minipage}
\hfill

% \end{tcolorbox}
\end{minipage}

\caption{\textbf{Memorization Example of Qwen2.5-Math-7B on AIME2024.}
In this case, the first 40\% of the original problem is used as the prompt, and the generation is performed under the \textit{\greedyonly} configuration.}
\label{fig:example_qwen2.5_math_7b_aime2024}
\end{figure*}

%
%%%%%%%%%%%%%%%%%%%%%%%%%%%%%%%%%%%%%%%%%%%%%%%%%%%%%%%%%%%%%%%%%%%%%%%%%%%%%%%%

%
% 图表：对比展示两个模型的输出
%
\begin{figure*}[!ht]
\centering
\begin{minipage}{0.9\textwidth}
% \begin{tcolorbox}[
%   enhanced,
%   attach boxed title to top center={yshift=-3mm},
%   colbacktitle=red!75!black,
%   colframe=gray!75!black,
%   colback=white,
%   width=\textwidth,
%   boxed title style={size=small, colback=red!75!black},
%   fonttitle=\bfseries\color{white},
%   title={}, %此处可以改标题
% ]

% 顶部框：题目部分
\begin{tcolorbox}[
  colback=white,
  colframe=MidnightBlue,
  coltitle=white,
  title=\textbf{Example from \textit{\randomcalculation}, 5 Calculation Steps},
  fonttitle=\bfseries
]
\textbf{Problem}: Evaluate this LaTeX numerical expression step-by-step and give the final value within \textbackslash boxed\{\}: 

\[
18^2/\frac{34}{8}+\frac{89}{4}-\frac{49}{9}*\frac{56}{4}+62^2
\]

\textbf{Answer}: \boxed{3866.26307189542}

\end{tcolorbox}

% 下方左右并排等高的两个框
\noindent
\begin{minipage}[t]{\textwidth}
  \begin{tcolorbox}[
    % colback=cyan!10, 
    % colframe=SkyBlue!80!black, 
    % coltitle=white,
    colback=teal!2!white, 
    colframe=teal!80!black, 
    coltitle=white,
    title=\textbf{Qwen2.5-Math-7B Output},
    fonttitle=\bfseries,
    height fixed for=all,
  ]
To evaluate the given LaTeX numerical expression step-by-step, we will follow the order of operations (PEMDAS/BODMAS): Parentheses/Brackets, Exponents/Orders, Multiplication and Division (from left to right), Addition and Subtraction (from left to right).

The expression is:
\[ 18^2 / \frac{34}{8} + \frac{89}{4} - \frac{49}{9} * \frac{56}{4} + 62^2 \]

Step 1: Evaluate the exponents.
\[ 18^2 = 324 \]
\[ 62^2 = 3844 \]
%So the expression becomes:
%\[ 324 / \frac{34}{8} + \frac{89}{4} - \frac{49}{9} * \frac{56}{4} + 3844 \]

...

Step 2: Simplify the fractions and perform the division.
\[ \frac{34}{8} = 4.25 \]
%So the expression becomes:
%\[ 324 / 4.25 + \frac{89}{4} - \frac{49}{9} * \frac{56}{4} + 3844 \]
%\[ 324 / 4.25 \approx 76.23529411764706 \]
%So the expression becomes:
%\[ 76.23529411764706 + \frac{89}{4} - \frac{49}{9} * \frac{56}{4} + 3844 \]
...

Step 3: Simplify the remaining fractions and perform the multiplication.
\[ \frac{89}{4} = 22.25 \]
\[ \frac{49}{9} \approx 5.444444444444444 \]
%\[ \frac{56}{4} = 14 \]
%\[ \frac{49}{9} * \frac{56}{4} \approx 5.444444444444444 * 14 \approx 76.22222222222223 \]

%So the expression becomes:
%\[ 76.23529411764706 + 22.25 - 76.22222222222223 + 3844 \]
...

Step 4: Perform the addition and subtraction from left to right.
\[ 76.23529411764706 + 22.25 \approx 98.48529411764706 \]
\[ 98.48529411764706 - 76.22222222222223 \approx 22.26307189542483 \]
\[ 22.26307189542483 + 3844 \approx 3866.263071895425 \]

So the final value of the expression is:
\[ \boxed{3866.263071895425} \]
  \end{tcolorbox}
\end{minipage}
\hfill

% \end{tcolorbox}
\end{minipage}

%
% RLVR在五步骤数据集上的正确输出示例。这里使用的模型是Qwen2.5-Math-7B在正确的奖励信号下，训练至500步。
%
\caption{Example of a correct output produced by RLVR on the 5-Step \textit{\randomcalculation} dataset. The result is obtained using the Qwen2.5-Math-7B model, trained for 300 steps under the correct reward signal.}
\label{fig:example_rlvr_qwen2.5_math_7b_5_step_correct}
\end{figure*}

%%%%%%%%%%%%%%%%%%%%%%%%%%%%%%%%%%%%%%%%%%%%%%%%%%%%%%%%%%%%%%%%%%%%%%%%%%%%%%%%
%
% 【代码评测】表格数据以及示例
\definecolor{Gray}{gray}{0.90}

\begin{table*}[t]
\centering
\small
\setlength{\tabcolsep}{4pt}

\begin{tabular}{lccccccc}
\toprule
\multirow{2}{*}{\textbf{Model}} 
& \multicolumn{2}{c}{\textbf{80\%-Problem}} 
& \multicolumn{2}{c}{\textbf{60\%-Problem}} 
& \multicolumn{2}{c}{\textbf{40\%-Problem}} \\
\cmidrule(lr){2-3} \cmidrule(lr){4-5} \cmidrule(lr){6-7}
& ROUGE-L & EM & ROUGE-L & EM & ROUGE-L & EM \\
\midrule
\textbf{Qwen2.5-Math-7B} 
& 83.47 & \textbf{56.59} & 75.34 & 35.16 & 66.66 & 20.33 \\

\textbf{Qwen2.5-Math-7B-Instruct} 
& 46.60 & 4.95 & 36.49 & 0.00 & 32.62 & 0.00 \\

\textbf{Qwen2.5-7B} 
& 94.54 & \textbf{85.71} & 93.43 & \textbf{81.32} & 91.46 & \textbf{70.88} \\

\textbf{Qwen2.5-7B-Instruct} 
& 77.50 & 48.90 & 70.79 & 26.37 & 67.03 & 16.48 \\

\textbf{Llama3.1-8B} 
& 45.28 & 4.40 & 37.53 & 1.10 & 29.78 & 0.00 \\

\textbf{Llama3.1-8B-Instruct} 
& 36.20 & 1.65 & 32.77 & 0.00 & 28.33 & 0.00 \\
\bottomrule
\end{tabular}
\caption{
Accuracy (Exact Match, EM) and ROUGE-L scores of different models on LiveCodeBench (test generation) under different prompt prefix ratios in greedy decoding mode (\textbf{\textit{\greedyonly}} configuration).
}
\label{tab:qwen_code_memorization_greedy}
\end{table*}
%
% 图表：对比展示两个模型的输出
%
\begin{figure*}[!t]
\centering
\begin{minipage}{0.9\textwidth}
% \begin{tcolorbox}[
%   enhanced,
%   attach boxed title to top center={yshift=-3mm},
%   colbacktitle=red!75!black,
%   colframe=gray!75!black,
%   colback=white,
%   width=\textwidth,
%   boxed title style={size=small, colback=red!75!black},
%   fonttitle=\bfseries\color{white},
%   title={}, %此处可以改标题
% ]

% 顶部框：题目部分
\begin{tcolorbox}[
  colback=gray!4!white,
  colframe=gray!85!black,
  coltitle=white,
  title=\textbf{Example from LiveCodeBench},
  fonttitle=\bfseries
]
\textbf{Problem}: You are given a 0-indexed integer array nums. You have to find the maximum sum of a pair of numbers from nums such that the maximum digit in both numbers are equal.
Return the maximum sum or -1 if no such pair exists.

\textbf{Test Cases}:\\
\textbf{Input}: [51, 71, 17, 24, 42]
\textbf{Output}: 88 \\
\textbf{Input}: [1, 2, 3, 4]
\textbf{Output}: -1

\end{tcolorbox}

\begin{tcolorbox}[
  colback=gray!4!white,
  colframe=gray!85!black,
  title=\textbf{Partial Prompt},
  fonttitle=\bfseries,
  boxrule=1pt,
  top=3mm,
  bottom=3mm
]
\textbf{Prompt}: You are given a 0-indexed integer array nums. You have to find the maximum sum of a pair of numbers from nums such that the maximum digit in both numbers are equal.
Return the maximum
\end{tcolorbox}

% 下方左右并排等高的两个框
\noindent
\begin{minipage}[t]{\textwidth}
  \begin{tcolorbox}[
    colback=teal!2!white, 
    colframe=teal!80!black, 
    coltitle=white,
    title=\textbf{Qwen2.5-Math-7B Output},
    fonttitle=\bfseries,
    height fixed for=all,
  ]
 sum or -1 if no such pair exists. \\
\textbf{Example 1:}\\
Input: nums = [51,71,17,24,42]
Output: 88\\
\textbf{Explanation:}
For i = 1 and j = 2, nums[i] and nums[j] have equal maximum digits with a pair sum of 71 + 17 = 88.
For i = 3 and j = 4, nums[i] and nums[j] have equal maximum digits with a pair sum of 24 + 42 = 66.
It can be shown that there are no other pairs with equal maximum digits, so the answer is 88.\\
\textbf{Example 2:}\\
Input: nums = [1,2,3,4]
Output: -1\\
\textbf{Explanation:} No pair exists in nums with equal maximum digits.
Constraints:
2 \textless= nums.length \textless= 100
1 \textless= nums[i] \textless= 104

...

\textbf{Code}
\begin{lstlisting}
class Solution {
    public int maxSum(int[] nums) {
        int max = -1;
        for(int i=0;i<nums.length-1;i++){
            for(int j=i+1;j<nums.length;j++){
                if(maxDigit(nums[i])==maxDigit(nums[j])){
                    max = Math.max(max,nums[i]+nums[j]);
                }
            }
        }
        return max;
    }
    public int maxDigit(int n){
        int max = 0;
        while(n>0){
            int rem = n%10;
            max = Math.max(max,rem);
            n = n/10;
        }
        return max;
    }
}
\end{lstlisting}
  \end{tcolorbox}
\end{minipage}
\hfill

% \end{tcolorbox}
\end{minipage}

\caption{\textbf{Memorization Example of Qwen2.5-Math-7B on LiveCodeBench.}
In this case, the first 80\% of the original problem is used as the prompt, and the generation is performed under the \textit{\greedyonly} configuration. Notably, the model even included the test cases in its output.}
\label{fig:example_qwen2.5_math_7b_livecodebench}
\end{figure*}

%%%%%%%%%%%%%%%%%%%%%%%%%%%%%%%%%%%%%%%%%%%%%%%%%%%%%%%%%%%%%%%%%%%%%%%%%%%%%%%%

\end{document}